\documentclass[10pt,twocolumn,letterpaper]{article}

\usepackage[pagenumbers]{cvpr} 

\usepackage{multirow}
\usepackage{float}
\newcommand{\best}[1]{\textbf{\textcolor{red}{#1}}}
\newcommand{\second}[1]{\textit{\textcolor{blue}{#1}}}

\usepackage{caption}
\usepackage{lineno}

\usepackage{wrapfig}

\usepackage{etoc}

\AddToShipoutPicture{%
  \AtTextUpperLeft{%
    \put(0,\LenToUnit{0.50in}){%
      \normalfont\fontsize{10}{12}\selectfont\color{black}%
      Published as a conference paper at CVPR 2026%
    }%
    \put(0,\LenToUnit{0.45in}){%
      \color{black}\rule{\textwidth}{0.4pt}%
    }%
  }%
}

\definecolor{cvprblue}{rgb}{0.21,0.49,0.74}
\usepackage[pagebackref,breaklinks,colorlinks,allcolors=cvprblue,hypertexnames=false]{hyperref}

\def\paperID{} 
\def\confName{CVPR}
\def\confYear{2026}

\title{AdaDexTrack: Dynamic Modulation for Adaptive and Generalizable Dexterous Manipulation Tracking}

\author{
Jianibieke Adalibieke$^{*1}$ \quad Qianwei Han$^{*1}$ \quad Xueyi Liu$^{2}$ \quad Yuzhe Qin$^{3}$ \quad Li Yi$^{\dagger2,1}$ \\ [0.3em]
$^{1}$Shanghai Qi Zhi Institute \quad $^{2}$Tsinghua University \quad $^{3}$UC San Diego \\ [0.5em]
{\tt\large \url{https://janebek.github.io/AdaDexTrack}} 
}

\begin{document}

\twocolumn[{%
\renewcommand\twocolumn[1][]{#1}%
\maketitle
\begin{center}
\vspace{-2.0em}
    \centering
    \captionsetup{type=figure,skip=1.8pt}
    \includegraphics[width=\textwidth]{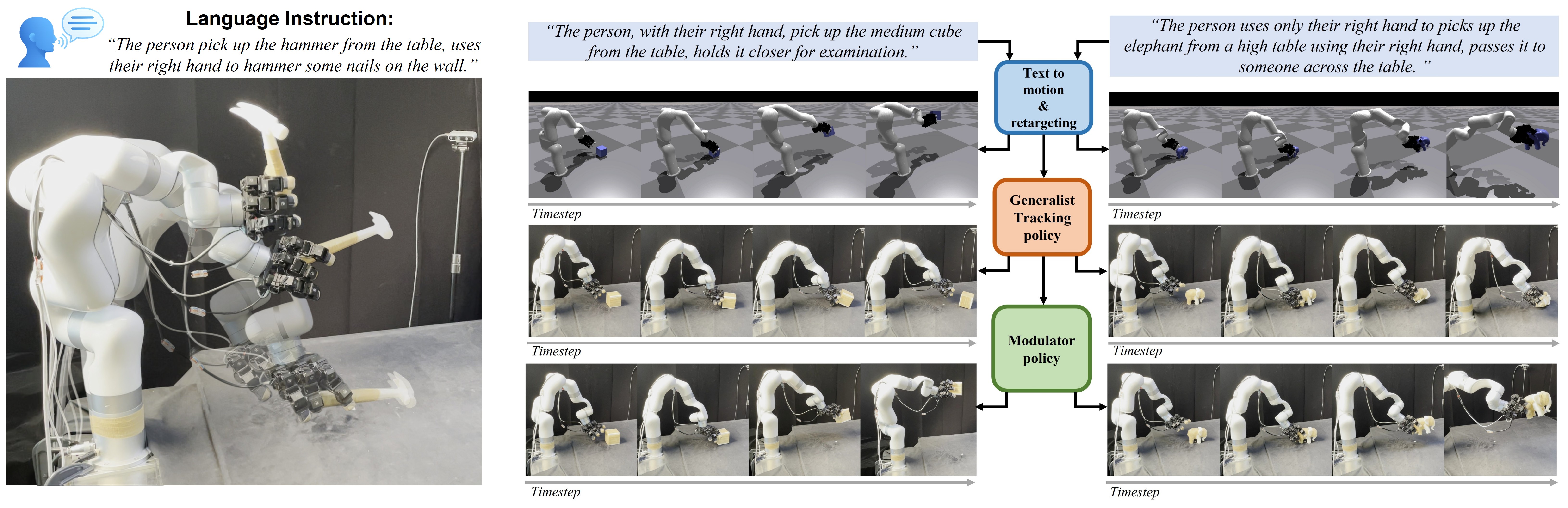}
\captionof{figure}{\textbf{AdaDexTrack for language-guided hand–object interaction tracking.} Given a noisy text-generated reference retargeted to the robot, the in-loop modulator enables robust long-horizon tracking. \textbf{Left:} successful reorientation of a thin object (Hammer). \textbf{Right:} generalization to an unseen object (CubeMedium) and an unseen trajectory (Elephant), where the tracker without the modulator fails—highlighting the benefit of in-the-loop modulation. Project website: \url{https://janebek.github.io/AdaDexTrack}}
\label{fig:teaserrrrrrrr}
\end{center}%
}]

\begin{abstract}

Language is a natural way to command robots, but converting a single instruction into a long-horizon, contact-rich hand–object interaction remains challenging: synthesized references are noisy, human-to-robot retargeting introduces embodiment bias, and fixed-reference tracking lets small errors snowball. We address this with AdaDexTrack, a modulator-in-the-loop framework for language-conditioned manipulation tracking. A distilled generalist tracker serves as the skill carrier, while a tightly aligned modulator performs three feedback corrections: reference modulation (continual adjustment of what to track), object-latent modulation (online adaptation of the object representation to recruit suitable skills), and positional-target modulation (small state-dependent refinements for execution). 
The tracker is learned via large-scale specialist to generalist distillation on a corpus of language-conditioned hand–object trajectories; the modulator is trained with RL under the same task objective, ensuring tight coupling. Across large-scale evaluations, AdaDexTrack consistently outperforms prior SOTA on unseen-trajectory and unseen-object sets in both average tracking error and success rate, demonstrating robustness and generalization. We further show zero-shot sim-to-real transfer on real hardware, where adding the modulator yields substantial gains over a tracker-only variant. AdaDexTrack reframes language-conditioned dexterous manipulation as modulated tracking, replacing the open-loop, fixed-reference tracking with in-loop modulation that adjusts the reference, object latent, and positional target, yielding drift-resistant execution from noisy text references.


\end{abstract}    
\section{Introduction}
\label{sec:intro}

Language is a natural interface for commanding robots. We study language-guided hand–object interaction (HOI) tracking: from an instruction and object geometry, a time-indexed hand–object reference is generated and then executed by a multi-fingered hand, requiring stable tracking across contact-rich, long-horizon behaviors.

An intuitive approach is a two-stage system: first, text-to-motion synthesizes an HOI trajectory from the instruction; second, interaction tracking executes that trajectory on the robot. Although both components have advanced, end-to-end coupling for text-guided manipulation tracking remains nontrivial and rarely demonstrated. Two challenges dominate: (i) noisy references—motion synthesis is imperfect and human-to-robot retargeting introduces embodiment bias; and (ii) open-loop planning under perfect-tracking assumptions—the tracker follows a fixed reference without adapting to how execution unfolds, so small errors accumulate over long horizons and the system eventually loses track. In practice, most trackers treat the reference as fixed~\cite{liu2025dextrack,li2025maniptrans}. When execution deviates, the controller is forced to chase it. Without conditioning on the robot’s current state or capabilities, this chasing degrades control. The policy becomes aggressive, contacts destabilize, and drift accelerates. We therefore advocate modulated tracking: dynamically adjusting the tracking reference, object representations and positional targets—to steer the controller back toward the original reference and maintain stable tracking. By putting these modulations in the feedback loop, the system corrects deviations on the fly and more reliably tracks long-horizon, text-guided trajectories.

We instantiate this idea with AdaDexTrack, a modulator-in-the-loop framework for language-conditioned manipulation tracking. AdaDexTrack has two components: (i) a distilled generalist tracker—the skill carrier—that encodes behaviors learned from diverse demonstrations; and (ii) a tightly aligned modulator that closes the loop via reference, object-latent, and positional-target modulation. Together, they turn noisy, language-conditioned references into drift-resistant executions. The general tracker is trained with large-scale teacher–student distillation: using DiffH2O~\cite{christen2024diffh2o} we synthesize text-guided hand–object trajectories, train a large set of teacher policies with reinforcement learning (RL) to track them, and distill their behaviors into a single general tracking policy. This policy has broad coverage but remains imperfect—exactly where the in-the-loop modulator adds value.

Based on the general tracker, the modulator adapts the tracking process in three ways
: (i) reference modulation—adding a small residual to the language-conditioned reference to continually re-align what to track and reduce distribution shift; (ii) object latent modulation—adjusts the object latent to invoke the tracker’s most suitable skills, composing behaviors best matched to the current reference; and (iii) positional target modulation—apply a small, state-dependent correction to the tracker’s positional targets to refine execution and quickly recover from drift.
The modulator shares the same task objective as the tracker and is trained with RL, ensuring tight objective alignment. Together, these modulations make tracking more robust to noise and embodiment mismatch, stabilize long-horizon execution, and improve generalization.


We conduct extensive evaluations in both the simulation and the real world to demonstrate the effectiveness of AdaDexTrack. In simulation, we evaluate on a large-scale text-guided HOI reference dataset and find that AdaDexTrack consistently outperforms all previous state-of-the-art (SOTA) methods on both unseen-trajectory and unseen-object splits in overall average tracking error and success rate, demonstrating robustness and generalization. In the real world, we achieve zero-shot sim-to-real transfer. We show that adding the modulator yields substantial gains over the tracker-only variant on both unseen-trajectory and unseen-object splits, highlighting the benefit of in-the-loop modulation.

To summarize, our contributions are: (i) a re-framing of language-conditioned dexterous manipulation tracking as modulated tracking, augmenting a strong but imperfect distilled tracker with an in-the-loop modulator; (ii) AdaDexTrack, a three-interface modulation scheme—reference, object-latent, and positional-target modulation—where the modulator shares the tracker’s task objective, and the tracker is learned via large-scale teacher–student distillation from synthesized text-guided HOI trajectories;
and (iii) extensive experiments and ablations showing consistent gains over prior SOTA on unseen-trajectory and unseen-object splits, with zero-shot sim-to-real transfer on hardware further demonstrating the effectiveness of the modulator-in-the-loop design.

\section{Related Works}
\label{sec:Related_Works}

\noindent\textbf{Dexterous Manipulation via HOI Reference.}
Learning from HOI references is a promising route to human-like dexterity, as such references capture contact timing, coordination synergies, and long-horizon intent. These references can come from (i) teleoperation, which offers high fidelity but is hard to scale~\cite{qin2023anyteleop,si2024tilde,ding2024bunny,zhang2025doglove,qin2022one,lin2025typetele,zhang2025kinedex,liu2024realdex}; (ii) mocap- or video-based human reconstruction, which is more scalable but labor-intensive, noisy, and subject to embodiment gaps~\cite{zhan2024oakink2,liu2022hoi4d,fan2023arctic,liu2024taco,taheri2020grab,banerjee2025hot3d}; or (iii) text-to-motion synthesis, which enables language-conditioned HOI generation at higher scale despite synthesis and retargeting noise~\cite{christen2024diffh2o,fu2025gigahands, peng2025hoi, wu2024thor, dai2024interfusion, cha2024text2hoi}. A practical way to use these motions is to represent them as time-indexed references and train a policy to track them. Prior RL/IL approaches often learn per-trajectory policies or generalize narrowly~\cite{yuan2025hermes,li2025maniptrans,shi2025learning,hsieh2025dexman,qin2022dexmv,chen2025vividex}. DexTrack~\cite{liu2025dextrack} broadens coverage by distilling multiple specialists into a single generalist tracker, but still treats tracking as a monolithic controller with a fixed test-time reference. This is fragile in contact-rich settings, where references are noisy and human-biased, embodiment and sensing mismatches introduce estimation error, and small deviations accumulate over long horizons. Classical trajectory modulation methods such as Dynamic Movement Primitives (DMP)~\cite{lu2024dynamic,saveriano2023dynamic,chen2023object,zhang2024logic} and Riemannian Motion Policies (RMP)~\cite{ratliff2018riemannian,van2022geometric,xie2023neural} can adapt motions, but typically rely on explicit hand-designed dynamical representations. We retain the tracking interface while treating the tracker as a distilled skill carrier and placing a modulator in the loop to adjust the reference, object latent, and positional target online, not to perfectly track the reference, but to correct synthesis noise in the HOI reference and accomplish the overall task.

\noindent\textbf{Skill Hierarchy in Dexterous Manipulation.}
In dexterous manipulation, a broad class of methods follows a skill hierarchy template: build a library of reusable skills and use a higher-level mechanism to select, sequence, or parameterize them for long-horizon dexterity. Representative instances include open-loop, hand-crafted primitives that compose into longer programs~\cite{bhatt2022surprisingly}, planning-/compliance-based decompositions without learning~\cite{morgan2022complex}, training-time guidance via simple sub-skill controllers with a learned switch~\cite{khandate2024dexterous}, low-level teleoperation skills to scale demonstration collection~\cite{lin2025learning}, multi-skill reuse where task transitions are user specified~\cite{gupta2021reset}, and learned transition-feasibility models that chain dexterous sub-policies and enable long-horizon switching/recovery~\cite{chen2023sequential}. ObjDex~\cite{chen2024object} fits this paradigm with an object-centric planner that proposes wrist trajectories tracked by a low-level controller. While effective, such stacks are loosely coupled: they act primarily via wrist-space proposals, optimize different objectives across levels, and keep the time-indexed reference and the tracker’s object representation outside the feedback loop—allowing systematic bias and embodiment mismatch to accumulate into drift. In contrast, we keep the tracking interface and add a tightly aligned in-loop modulator with three channels—reference modulation, object latent modulation, and positional target modulation—yielding more stable, long-horizon tracking.

\section{Method}

\begin{figure*}[h]
  \centering
  \vspace{-5pt}
  \includegraphics[width=\textwidth]{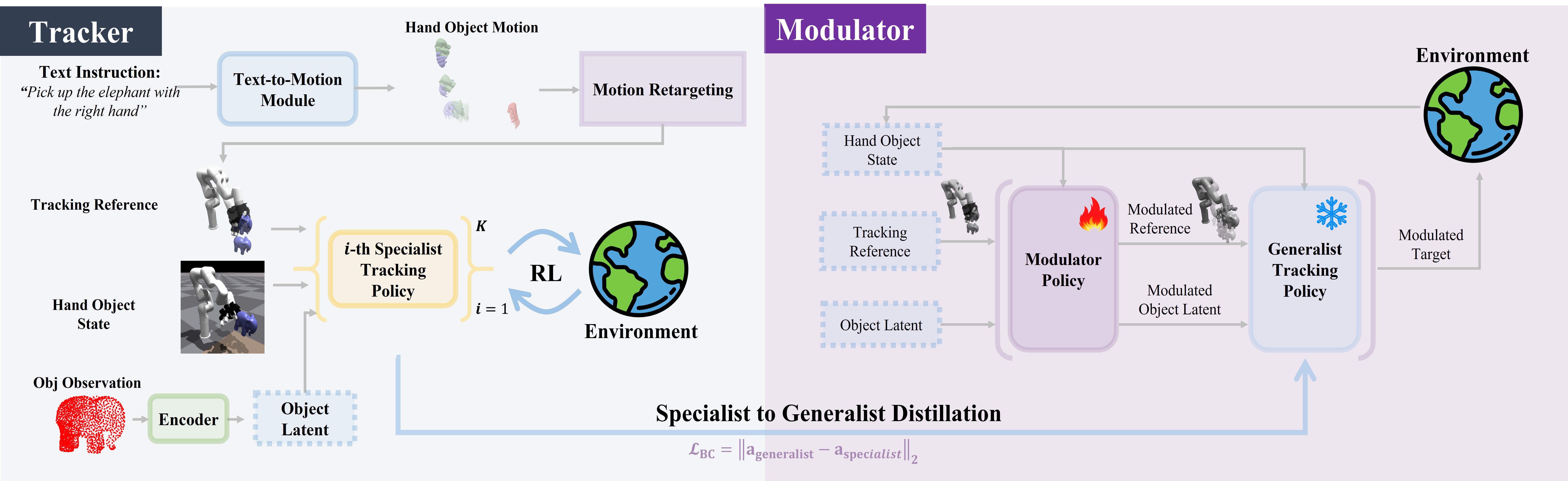}
  \vspace{-13pt}
  \caption{\textbf{AdaDexTrack pipeline overview.} \textbf{Left:} From a large-scale set of text-guided HOI references, we train many RL teacher trackers and distill their behaviors into a single generalist tracker. \textbf{Right:} On top of this tracker, we insert a modulator in the loop that performs reference modulation, object latent modulation, and positional target modulation, yielding drift-resistant, accurate long-horizon tracking.}

  \vspace{-12pt}
  \label{fig_method_pipeline}
\end{figure*}

\noindent\textbf{Overview.} 
Fixed text-generated references, together with embodiment mismatch, lead to long-horizon tracking drift; instead of forcing a single controller to track a fixed reference, we close the loop by modulating what the tracker sees and executes.
Fig.~\ref{fig_method_pipeline} gives an overview.
We split the framework into two stages:
(i) a general dexterous manipulation tracking policy $\pi^{\text{track}}$ that serves as the tracking controller and follows the HOI reference; and
(ii) a modulation policy $\pi^{\text{modulate}}$ that acts as a tightly aligned modulator, performing reference, object-latent, and positional-target modulation to maximize the tracker’s performance under feedback.
We first describe how we synthesize a large-scale, text-guided HOI reference dataset and develop a generalizable tracking controller from it (Sec.~\ref{sec:tracking}), then introduce the modulator and its integration with the tracker (Sec.~\ref{sec:modulator}), and finally present sim-to-real details (Sec.~\ref{sec:sim2real}).


\noindent\textbf{Task Formulation.} From a raw human–object trajectory $\{\hat{\boldsymbol{h}}_t\}_{t=1}^T$, we retarget to a robotic hand kinematic sequence $\{\hat{\boldsymbol{s}}_t\}_{t=1}^T$ and pose dexterous manipulation tracking as mimicking this sequence. A control policy is learned to mimic the kinematic demonstration while compensating for retargeting errors and motion-generator noise. 
At each timestep, the reference hand state is defined as $\hat{\boldsymbol{s}}_t^{h}
= \bigl(\hat{\boldsymbol{q}}_t^{h},\, \hat{\boldsymbol{p}}_t^{h}\bigr)$, where $\hat{\boldsymbol{q}}_t^{\,h} \in \mathbb{R}^{\,22}$ denotes the joint positions of the arm–hand chain, and $\hat{\boldsymbol{p}}_t^{\,h} \in \mathbb{R}^{\,5\times 7}$ stacks the rigid-body poses of the tracked links. Specifically, the robotic hand has 16 joints and the robotic arm has 6 joints, and the five tracked links are the four fingertips and the wrist of the robotic hand. The reference object state is defined as $\hat{\boldsymbol{s}}_t^{o}
= \hat{\boldsymbol{p}}_t^{o}$, where $\hat{\boldsymbol{p}}_t^{\,o} \in \mathbb{R}^{\, 7}$ is the object pose. 

We then formulate the dexterous manipulation tracking task as a Markov Decision Process (MDP), $\mathcal{M} = (\boldsymbol{\mathcal{S}}, \boldsymbol{\mathcal{A}}, \pi, \boldsymbol{\mathcal{T}}, \boldsymbol{\mathcal{R}}, \gamma)$, where $\boldsymbol{\mathcal{S}}$ is the state space, $\boldsymbol{\mathcal{A}}$ is the action space, $\pi$ is the policy, $\boldsymbol{\mathcal{T}}$ is the transition dynamics, $\boldsymbol{\mathcal{R}}$ is the reward function, $\gamma$ is the discount factor. At time step $t$, the policy maps the reference and current hand--object states to an action, $\boldsymbol{a}_t=\pi(\hat{\boldsymbol{s}}_t,\,\boldsymbol{s}_t)$, where $\boldsymbol{a}_t\in\mathbb{R}^{22}$ denotes a joint-space action and is controlled by a joint-space proportional--derivative (PD) controller.

\subsection{Tracking Controller}
\label{sec:tracking}
Training a generalizable tracker is hard under diverse, noisy language-conditioned references. We address this by (i) synthesizing a large-scale, text-guided HOI reference dataset; (ii) training specialist policies with RL to track these references; and (iii) distilling a large set of RL-trained specialists into a single policy capable of robustly tracking diverse HOI references.

\noindent\textbf{Language-Guided HOI Reference Generation.} One of the key challenges in learning a generalizable tracking controller is the scale of the reference dataset. To improve dataset scalability, we leverage the off-the-shelf motion-synthesis method DiffH2O~\cite{christen2024diffh2o} to generate dexterous hand–object interaction trajectories from natural-language prompts. Specifically, we leverage the detailed textual annotations provided by DiffH2O and use GPT-5~\cite{openai_chatgpt5_2025} for paraphrase augmentation: for each annotated trajectory, we generate multiple semantically equivalent descriptions—following DiffH2O’s native annotation format—to increase linguistic variety while preserving intent. We then feed these paraphrases back into DiffH2O to synthesize additional demonstrations. This produces a dataset that is diverse yet plausible: the generated descriptions conform to DiffH2O’s annotation style and schema, which keeps the synthesized motions within DiffH2O’s generative distribution and reduces distribution shift.

\noindent\textbf{Specialist Policy Training.} Based on the human--object references generated by DiffH2O, $\{\hat{\boldsymbol{h}}_n\}_{n=1}^{N}$, we first retarget these motions to the robot arm--hand kinematic chain, yielding $\{\hat{\boldsymbol{s}}_n\}_{n=1}^{N}$. For retargeting,  we manually define correspondences between the human-hand and robot-hand meshes and minimize the distances between corresponding points, in a manner similar to DexTrack~\cite{liu2025dextrack}. For each kinematic reference $\hat{\boldsymbol{s}}_{i}$, we train a specialized tracking controller $\pi^{\text{track}}_{i}$ to minimize the discrepancy between the robot state and the reference. We formulate this as a reinforcement learning (RL) problem and train each $\pi^{\text{track}}_{i}$ with Proximal Policy Optimization (PPO)~\cite{schulman2017proximal}. For notational simplicity, we drop the subscript \(i\) and write \(\pi^{\text{track}}\) and \(\hat{\boldsymbol{s}}\); unless otherwise stated, they refer to a fixed but arbitrary \(i\).

At timestep $t$, the observation for $\pi^{\text{track}}$ is defined as $\boldsymbol{o}_t^{\text{track}}=\bigl(\boldsymbol{s}_t^{\text{track}}, \hat{\boldsymbol{s}}_t^{\text{track}}\bigr)$, where the current state $\boldsymbol{s}_t^{\text{track}}=\bigl(\boldsymbol{s}_t^{\text{prop}}, \boldsymbol{s}_t^{\text{o}}, \boldsymbol{f}_t^{\text{o}}\bigr)$ concatenates robot proprioception $\boldsymbol{s}_t^{\text{prop}}$, the object pose $\boldsymbol{s}_t^{\text{o}}$, and an object latent $\boldsymbol{f}^{\text{o}}$. The latent is produced by a pre-trained point-cloud encoder $E_{\text{pc}}$ applied to the object point cloud $\mathcal{P}$, i.e., $\boldsymbol{f}^{\text{o}}=E_{\text{pc}}(\mathcal{P})$. The reference state is defined as $\hat{\boldsymbol{s}}_{t}^{\text{track}}=\bigl(\hat{\boldsymbol{s}}_{t}^{\text{h}},\,\hat{\boldsymbol{s}}_{t}^{\text{o}}\bigr)$, where $\hat{\boldsymbol{s}}_{t}^{\text{h}}$ and $\hat{\boldsymbol{s}}_{t}^{\text{o}}$ denote the reference hand and object states, respectively. We include not only the current reference $\bigl(\hat{\boldsymbol{s}}_{t}^{\text{h}},\,\hat{\boldsymbol{s}}_{t}^{\text{o}}\bigr)$ but also future references $\bigl(\hat{\boldsymbol{s}}_{t+k}^{\text{h}},\,\hat{\boldsymbol{s}}_{t+k}^{\text{o}}\bigr)$ as observations, with $k\in\{1,2,4,12\}$. All terms are concatenated and treated as the tracking reference state:
$\hat{\boldsymbol{s}}_{t}^{\text{track}}=\operatorname{concat}_{k\in\{0,1,2,4,12\}}\!\big(\hat{\boldsymbol{s}}_{t+k}^{\text{h}},\,\hat{\boldsymbol{s}}_{t+k}^{\text{o}}\big)$. For the action space, the tracking policy $\pi^{\text{track}}$ takes the observation $\boldsymbol{o}_t^{\text{track}}$ as input and outputs an action $\boldsymbol{a}_t \in \mathbb{R}^{22}$ consisting of 22-D relative joint position changes for the hand and arm revolute joints. The reward function is designed to align the transited hand and object states with their respective references and to encourage hand–object proximity~\cite{wan2023unidexgrasp++}. Formally, it is defined as:
\begin{equation}\label{eq:rt-loss}
\begin{aligned}
r_t ={}& \omega_1 \lVert \boldsymbol{q}_t^{\mathrm h}-\hat{\boldsymbol{q}}_t^{\mathrm h}\rVert_2^2
      + \omega_2 \lVert \boldsymbol{p}_t^{\mathrm h}-\hat{\boldsymbol{p}}_t^{\mathrm h}\rVert_2^2 \\
    &+ \omega_3 \lVert \boldsymbol{p}_t^{\mathrm o}-\hat{\boldsymbol{p}}_t^{\mathrm o}\rVert_2^2
      + \omega_4 \lVert \boldsymbol{p}_t^{\mathrm h}-\boldsymbol{p}_t^{\mathrm o}\rVert_2^2
\end{aligned}
\end{equation}

\noindent In Eq.~\eqref{eq:rt-loss}, $\lVert \boldsymbol{q}_t^{\mathrm{h}}-\hat{\boldsymbol{q}}_t^{\mathrm{h}}\rVert_2^2$ is the joint-position tracking cost for the robot arm--hand chain, where $\boldsymbol{q}_t^{\mathrm{h}}$ and $\hat{\boldsymbol{q}}_t^{\mathrm{h}}$ are the current and reference joint positions, respectively; $\lVert \boldsymbol{p}_t^{\mathrm{h}}-\hat{\boldsymbol{p}}_t^{\mathrm{h}}\rVert_2^2$ is the rigid-body tracking cost for the fingers and wrist, where $\boldsymbol{p}_t^{\mathrm{h}}$ and $\hat{\boldsymbol{p}}_t^{\mathrm{h}}$ are the current and reference rigid-body poses; $\lVert \boldsymbol{p}_t^{\mathrm{o}}-\hat{\boldsymbol{p}}_t^{\mathrm{o}}\rVert_2^2$ is the object-pose tracking cost, where $\boldsymbol{p}_t^{\mathrm{o}}$ and $\hat{\boldsymbol{p}}_t^{\mathrm{o}}$ are the current and reference object poses; and $\lVert \boldsymbol{p}_t^{\mathrm{h}}-\boldsymbol{p}_t^{\mathrm{o}}\rVert_2^2$ is a proximity term that encourages a small hand--object distance. The coefficients $\omega_1,\omega_2,\omega_3,\omega_4$ weight the corresponding terms and are chosen to balance the reward. Please see the supplementary material for details on the observation and action spaces and the reward.

\noindent\textbf{Generalist Distillation.} To distill knowledge from multiple specialists into a single general policy $\pi_{\text{general}}^{\text{track}}$, we first collect a set of specialist policies $\{\pi_i^{\text{track}}\}_{i=1}^N$. For each annotated reference trajectory and its generated semantically equivalent trajectories, we train a specialized specialist $\pi_i^{\text{track}}$ to track them. This specialization constrains each specialist to a narrow, semantically consistent behavior, thereby reducing the RL exploration space and accelerating training. Given $\{\pi_i^{\text{track}}\}_{i=1}^N$, we construct a dataset $D$ of successful trajectories generated by the specialists, $D=\bigcup_{i=1}^N \{\tau^{(i)}_m\}_{m=1}^{M_i}$, where each $\tau^{(i)}_m$ is a successful trajectory sampled from $\pi_i^{\text{track}}$. We then train the general policy $\pi_{\text{general}}^{\text{track}}$ via Behavior Cloning (BC)~\cite{yuan2024robot,liu2025dexndm}. Although DAgger-based distillation has proven effective in previous work with a small number of specialists, it requires multiple rounds of interaction with the environment, where the generalist queries specialists for action labels at each visited state. However, with a large set of specialists in our case, this results in substantial increases in both training time and computational cost, making the offline BC approach more efficient in our setting. The observation space of the general policy is the same as that of the specialist policy, $\boldsymbol{o}_t^{\text{track}}$, enabling direct real-world deployment without oracle states.

\subsection{Tracking Modulator}
\label{sec:modulator}
The general tracking policy $\pi_{\text{general}}^{\text{track}}$ is trained using a large set of specialist tracking policies via BC. Next, we organize the controller as a hierarchical policy: an RL-based modulator $\pi^{\text{modulate}}$ is placed before the tracker $\pi^{\text{track}}$. It conditions the tracker in three ways (detailed below) to enhance generalization and reliability.

\noindent\textbf{Reference Modulation.} Because the tracking controller is imperfect---long-horizon tracking cannot perfectly follow the reference---errors accumulate over time, causing drift and out-of-distribution (OOD) states. To mitigate this, we exploit the tracker's short-horizon accuracy and dynamically advance the target by using a near-future hand reference $\hat{\boldsymbol{s}}_{t+k}$ with $k\in\{1,2,4,12\}$. Specifically, at each step $t$, we update the near-future reference with a unified rule, $\hat{\boldsymbol{s}}^{x}_{t+k}\leftarrow \hat{\boldsymbol{s}}^{x}_{t+k}+\lambda_{x}\boldsymbol{a}^{\mathrm{ref},x}_{t+k}$ for $k\in\{1,2,4,12\}$ and $x\in\{\mathrm{h},\mathrm{o}\}$ ($x=\mathrm{h}$: hand; $x=\mathrm{o}$: object), where $\boldsymbol{a}^{\mathrm{ref},x}_{t+k}$ is the modulator action used to modulate the reference and $\lambda_x$ controls the adjustment magnitude to keep the modified reference close to the original trajectory. By continuously updating the reference at each timestep $t$, the modulator leverages the tracker's strong short-horizon accuracy, thereby canceling accumulated error and reducing long-horizon drift and overall tracking error.

\noindent\textbf{Object Latent Modulation.} For the general tracker, the observation is defined as $\boldsymbol{o}_t^{\text{track}}=\bigl(\boldsymbol{s}_t^{\text{track}},\,\hat{\boldsymbol{s}}_t^{\text{track}}\bigr)$, where the current state $\boldsymbol{s}_t^{\text{track}}=\bigl(\boldsymbol{s}_t^{\text{prop}},\,\boldsymbol{s}_t^{\text{o}},\,\boldsymbol{f}^{\text{o}}\bigr)$ concatenates robot proprioception $\boldsymbol{s}_t^{\text{prop}}$, the estimated object pose $\boldsymbol{s}_t^{\text{o}}$, and the object latent $\boldsymbol{f}^{\text{o}}$. During distillation, we log $(\boldsymbol{o}_t^{\text{track}},\boldsymbol{a}_t)$ pairs from each specialist; with $\boldsymbol{f}^{\text{o}}$ explicitly embedded in $\boldsymbol{o}_t^{\text{track}}$, these data instantiate a skill library of state--action primitives conditioned on the object--latent subspace. At inference time, the general tracker computes a fixed $\boldsymbol{f}^{\text{o}}$ at task start and keeps it constant throughout execution, acting as
$\boldsymbol{a}_t=\pi_{\text{track}}\!\bigl(\boldsymbol{s}_t^{\text{prop}},\,\boldsymbol{s}_t^{\text{o}},\,\boldsymbol{f}^{\text{o}},\,\hat{\boldsymbol{s}}_t^{\text{track}}\bigr)$.
 Our modulator relaxes this restriction by producing a modulated latent $\tilde{\boldsymbol{f}}_t^{\text{o}}$ and forming the modulated observation $\tilde{\boldsymbol{o}}_t^{\text{track}}=\bigl(\tilde{\boldsymbol{s}}_t^{\text{track}},\,\hat{\boldsymbol{s}}_t^{\text{track}}\bigr)$ with $\tilde{\boldsymbol{s}}_t^{\text{track}}=\bigl(\boldsymbol{s}_t^{\text{prop}},\,\boldsymbol{s}_t^{\text{o}},\,\tilde{\boldsymbol{f}}_t^{\text{o}}\bigr)$. Although trained with BC, the general tracker learns a continuous mapping from $(\boldsymbol{o}_t^{\text{track}},\boldsymbol{f}_t^{\text{o}})$ to $\boldsymbol{a}_t$. As $\boldsymbol{f}_t^{\text{o}}$ changes, the general tracker's behavior moves smoothly between specialists, allowing the modulator’s object latent modulation to recruit the most suitable primitives at runtime. While the skill library is built from a finite set of object latents $\{\boldsymbol{f}^{\text{o}}_i\}_{i=1}^{M}$, the modulator treats $\tilde{\boldsymbol{f}}^{\text{o}}_t\!\in\!\mathbb{R}^{64}$ as a continuous control variable, effectively turning a discrete ability menu into a continuous manifold that can interpolate between and switch across object anchors.

\noindent\textbf{Positional Target Modulation.} 
We apply a small, bounded action space correction to absorb execution-level discrepancies that the tracker cannot capture alone. Given the tracker’s command $\boldsymbol{a}_t=\pi_{\text{track}}(\boldsymbol{o}_t^{\text{track}})$, the modulator outputs a residual positional target $\Delta\boldsymbol{a}_t$, and the final command is
$\boldsymbol{a}'_t=\boldsymbol{a}_t+\Delta\boldsymbol{a}_t$.
The residual targets fast, local, high-frequency nonidealities, providing on-the-fly refinement while keeping the tracker in charge. Intuitively, $\boldsymbol{a}_t$ carries the low-frequency structure of the skill learned during distillation, whereas $\Delta\boldsymbol{a}_t$ supplies a small and smooth compensation that improves precision and robustness.



\subsection{Sim-to-Real Transfer}
\label{sec:sim2real}
To reduce the sim-to-real gap, we perform system identification to tune the simulator's dynamics parameters to best match the real-world system, and apply domain randomization in simulation to span real-world physics variability, thereby improving robustness and transfer.

\noindent\textbf{System Identification.} 
The dynamics parameters we identify in simulation are the joint stiffness $P$ and joint damping $D$. We first collect real-world joint state--action trajectories by actuating the real robot with sinusoidal action commands at multiple frequencies and amplitudes. 
After collecting real-world data, we use the black-box optimizer CMA-ES~\cite{hansen2003reducing} to identify joint stiffness $P$ and joint damping $D$ that best match the real robot’s motion. We estimate $(P^\star,D^\star)$ by minimizing the discrepancy between simulated and real joint states, with $L=\sum_{m=1}^{M}\sum_{t=1}^{N}\|\boldsymbol{q}_{m,t}^{\,s}-\boldsymbol{q}_{m,t}^{\,r}\|_2^2$ and $(P^\star,D^\star)=\arg\min_{P,D} L$, where $M$ is the number of trajectories and $N$ is the length of the trajectory. Please see the supplementary material for additional details.

\noindent\textbf{Domain Randomization.} Based on the system identification results, we apply small randomization to robot dynamics parameters (link masses, joint friction, joint stiffness, joint damping) and large randomization to object dynamics (mass, friction, restitution). We also inject Gaussian noise into the robot's proprioceptive states and action commands. In addition, we randomize the object's initial pose relative to the object's initial reference state $\hat{\boldsymbol{s}}_{0}^{\mathrm{o}}$. Finally, we apply high-friction tape to the fingertips and the object, as we empirically observe that higher friction yields higher success rates and less failures.
\section{Experiments}
We first describe the experimental setup (Sec.~\ref{sec:setup}). We then compare our approach with SOTA baselines, demonstrating its generalization and effectiveness (Sec.~\ref{sec:main}). Next, we conduct ablations of the modulator, isolating the effects of modulating the reference and the object latent, and further analyze how the amount of training data impacts generalization (Sec.~\ref{sec:ablation}). Finally, we present closed-loop zero-shot sim-to-real results on real hardware (Sec.~\ref{sec:real}).


\begin{table*}[!t]
\centering
\small
\setlength{\tabcolsep}{6pt}
\setlength{\abovecaptionskip}{1.0pt}   
\begin{tabular}{@{}c c c c c c c c@{}}
\toprule
Dataset & Method &
$T_{\mathrm{o}}$ (cm, $\downarrow$) &
$R_{\mathrm{o}}$ (rad, $\downarrow$) &
$T_{\mathrm{h}}$ (cm, $\downarrow$) &
$R_{\mathrm{h}}$ (rad, $\downarrow$) &
$E_{\mathrm{finger}}$ (rad, $\downarrow$) &
Succ. (\%, $\uparrow$)\\
\midrule
\multirow{6}{*}{\parbox{1.2cm}{\centering Unseen\\Trajectory}} 
& ObjDex          & 4.95 & 0.6950 & \second{8.02}            & 0.6296      & 0.6872      & 51.72/44.59 \\
& DexTrack                         & 6.93 & 0.7046
 & 9.17            & 0.5742
     & 0.2823
       & 76.13/73.89 \\
& Vanilla RL (PPO) & 12.10 & 0.8992
 & 13.30    & 0.9142    & 0.3595        & 43.03/53.94 \\
\cmidrule(lr){2-2} 
\cmidrule(lr){3-8} 
& Ours (General Tracker) & 11.57 & 0.7607
 & 10.33            & 0.5501   & 0.2791 & 62.20/65.01\\
& Ours (General Tracker+R)   & 10.06 & 0.7445 & 9.76 & \second{0.5408} & \second{0.2702}           & 67.09/68.14 \\
& Ours (General Tracker+R+T)                           & \second{4.68}  & \best{0.6272}  & \best{7.99} & \best{0.5062} & \best{0.2413} & \second{84.45}/\best{80.42}\\
& Ours (General Tracker+R+O+T)  & \best{4.49} & \second{0.6720}
 & 8.77    & 0.5860        & 0.2714   & \best{88.99}/\second{77.92} \\
\midrule
\multirow{6}{*}{\parbox{1.2cm}{\centering Unseen\\Object}} 
& ObjDex          & 20.29 & 1.0556 & 11.24            & 0.9483      & 0.6833      & 28.16/37.82 \\
& DexTrack                         & \second{18.06} & 0.9043
 & 10.32            & 0.6908
     & 0.2953
       & \second{39.59}/47.37 \\
& Vanilla RL (PPO) & 23.20 & \best{0.7830}
 & \second{9.63}    & \best{0.5828}    & \best{0.2533}        & 25.71/44.24 \\
\cmidrule(lr){2-2} 
\cmidrule(lr){3-8} 
& Ours (General Tracker) & 21.74 & 0.9562
 & 11.62            & \second{0.5964}   & 0.2845 & 24.08/42.52\\
& Ours (General Tracker+R)   & 22.26 & 0.9415 & 11.70 & 0.5995 & 0.2785           & 25.71/42.89 \\
& Ours (General Tracker+R+T)                           & 18.26  & \second{0.8966}  & 11.72 & 0.6819 & \second{0.2760} & 37.55/\second{47.51}\\
& Ours (General Tracker+R+O+T)  & \best{15.45} & 0.9492
 & \best{9.21}    & 0.6414        & 0.2922   & \best{46.12}/\best{53.78} \\
\bottomrule
\end{tabular}
\caption{Quantitative results. Best and second-best entries are highlighted in \textbf{\textcolor{red}{bold red}} and \textit{\textcolor{blue}{italic blue}}, respectively. Ablations are denoted as ``Ours (General Tracker+R)'', ``Ours (General Tracker+R+T)'', ``Ours (General Tracker+R+O+T)'', where R = reference modulation, O = object latent modulation, and T = positional target modulation. Success rate reported as Mean-error/Completion.} 
\label{tab:main}
\vspace{-0.7em}
\end{table*}


\subsection{Experiment Setup}
\label{sec:setup} 
\noindent\textbf{Datasets.}
As described in Sec.~\ref{sec:tracking}, we use DiffH2O~\cite{christen2024diffh2o} to generate human–object motion references. Leveraging the dataset’s text annotations, we filter the 1{,}333 annotated instances to retain only sequences that (i) involve single-hand (predominantly right-hand) manipulation with no bimanual interaction, (ii) exhibit no hand–robot-arm collisions, and (iii) keep the hand pose within the manipulator’s reachable workspace at every timestep; this yields 505 reference sequences. We then augment each retained sequence by generating nine semantically equivalent texts and synthesizing a motion reference from each, expanding the set to 5{,}050 sequences. Because the generative process can produce invalid references, we reapply the same feasibility criteria to the augmented set, yielding 2{,}765 valid sequences spanning 50 objects. We randomly partition these sequences into 2{,}212 for training and 553 for testing—an unseen-trajectory split. At the object level, we additionally form an unseen-object split by using 45 objects (2{,}520 sequences) for training and holding out 5 objects (245 sequences) for testing.

\noindent\textbf{Metrics.}
We evaluate tracking with six metrics: (1) \(T_{\mathrm{o}}\) (Object Position Error), the per-frame mean Euclidean distance between the reference and the tracked object positions; (2) \(R_{\mathrm{o}}\) (Object Rotation Error), the per-frame mean angular difference (degrees) between the reference and the tracked object orientations; (3) \(T_{\mathrm{h}}\) (Wrist Position Error), the per-frame mean Euclidean distance between the reference and the tracked wrist positions; (4) \(R_{\mathrm{h}}\) (Wrist Rotation Error), the per-frame mean angular difference (degrees) between the reference and the tracked wrist orientations; (5) \(E_{\text{finger}}\) (Finger Joint Error), the per-frame mean absolute joint-position error across finger DoFs; and (6) Success Rate (two variants):
Mean-error—the fraction of trajectories whose mean errors satisfy \(0.5\,T_{\mathrm{o}}+0.5\,T_{\mathrm{h}}\le 10\,\mathrm{cm}\), \(0.5\,R_{\mathrm{o}}+0.5\,R_{\mathrm{h}}\le 60^\circ\) and \(E_{\text{finger}}\le 40^\circ\); and Completion—for each trajectory, the fraction of frames \(t\) with \(0.5\,T_{\mathrm{o}}(t)+0.5\,T_{\mathrm{h}}(t)\le 10\,\mathrm{cm}\), \(0.5\,R_{\mathrm{o}}(t)+0.5\,R_{\mathrm{h}}(t)\le 60^\circ\) and \(E_{\text{finger}}(t)\le 40^\circ\) (the average over trajectories).

\noindent\textbf{Baselines.}
We compare our approach with three baselines. (1) DexTrack~\cite{liu2025dextrack}: a single generalist neural tracker distilled from many human–object demonstrations, trained with reinforcement learning and imitation learning (IL). Given the current state and a fixed reference, it outputs tracking actions directly, with no modulator in the loop. Since DexTrack’s homotopy-based demo mining and IL+RL training code are unavailable, we re-implemented the IL+RL scheme described in the paper: we execute the teacher on each reference to record its actions, and add a supervised imitation term to the actor’s RL objective as a regularizer. Homotopy-based mining is not included in our reproduction. (2) ObjDex~\cite{chen2024object}: a hierarchical planner-over-controller method learned from human hand motion data. A high-level, object-centric trajectory generator (planner) synthesizes wrist trajectories conditioned on the desired object goal, and a low-level RL controller tracks these trajectories with a dexterous hand. Since the high-level planner code is unavailable, we re-implemented the planner following the paper’s description and adapted the released RL controller and public configurations to our setting, retraining all components under our experimental protocol. (3) Vanilla RL (PPO): a policy trained with PPO~\cite{schulman2017proximal} instead of behavior cloning—using the same reference corpus and task reward—and with no modulator in the loop.

\subsection{Generalizable Tracking via Adaptation}
\label{sec:main}
We conduct experiments in Isaac Gym~\cite{makoviychuk2021isaac}. Table~\ref{tab:main} shows that AdaDexTrack attains the highest success rates on both unseen-trajectory and unseen-object sets, consistently surpassing all baselines. Compared to a monolithic tracker (DexTrack), which treats the reference as fixed and cannot adapt when synthesis/retargeting noise or embodiment mismatch pushes execution off-script, our in-loop modulator continually adjusts the reference, object latent, and applies a small, bounded positional target correction—preventing long-horizon error accumulation. Compared to a loosely coupled planner-over-controller (ObjDex), which corrects mainly in action/wrist space and leaves the reference and object encoding outside feedback, AdaDexTrack adapts at all three levels and remains tightly aligned with the tracker’s objective, yielding more stable contacts, smoother phase transitions, and better disturbance recovery. Ablations confirm that adding modulation channels improves both mean error and completion, indicating that the gains come from adaptation rather than model scale alone.


\subsection{Ablation Experiments}
\label{sec:ablation}
\noindent\textbf{Effect of Modulator Components.}
We ablate the modulator across its three components reference (R), object latent (O), and positional target (T). We evaluate ablations per split using three variants of our model: Ours (General Tracker+R), Ours (General Tracker+R+T), and Ours (General Tracker+R+O+T). We report results on both the unseen-trajectory and unseen-object sets.
As shown in Table~\ref{tab:main}, on both splits, increasing the number of modulator components yields consistent improvements in tracking performance.

\begin{figure}[!htbp]
  \centering
  \vspace{-5pt}
  \includegraphics[width=0.8\linewidth]{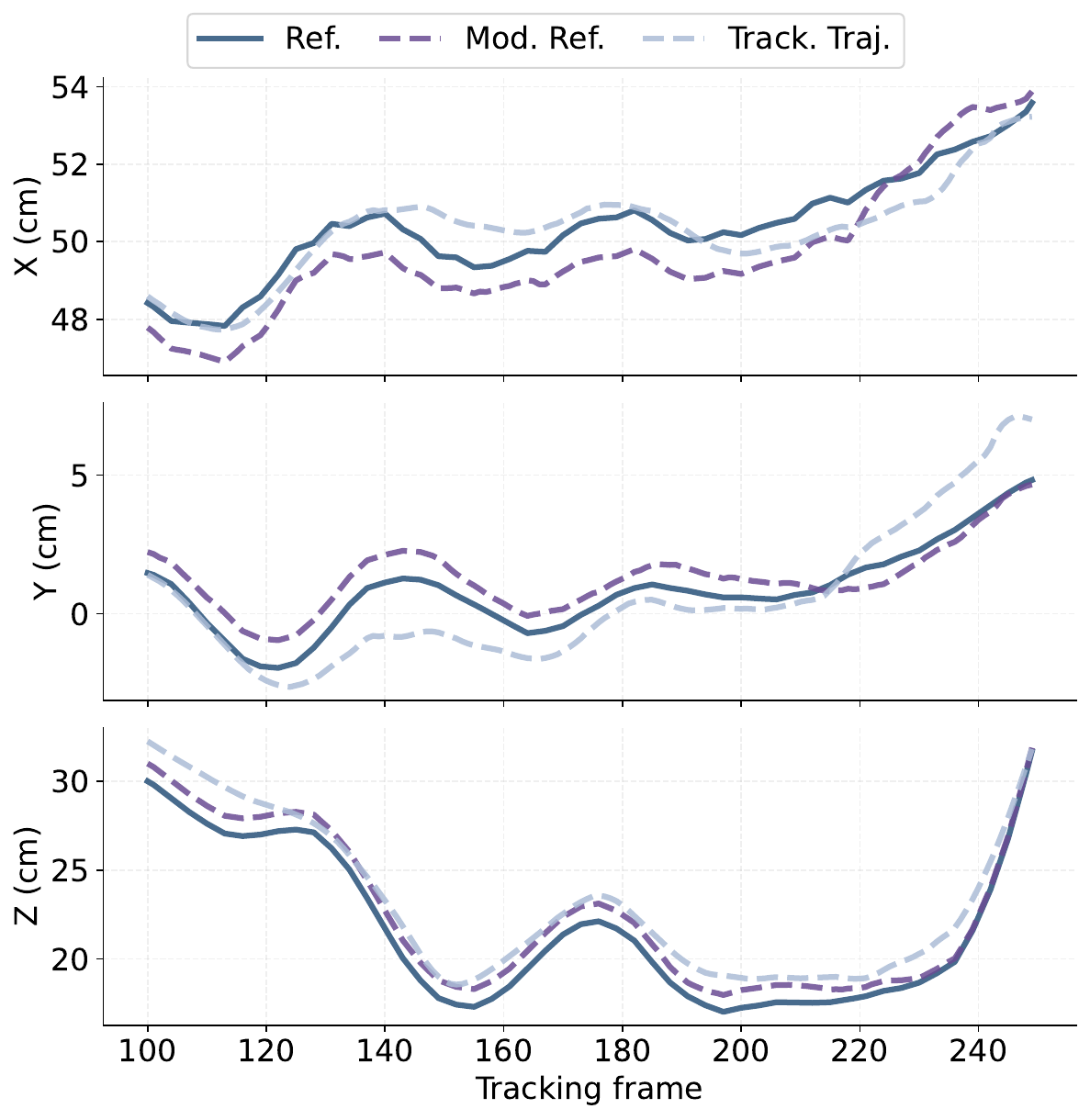}
  \vspace{-5pt}
  \caption{Reference modulation (General Tracker+R). \textbf{Ref.}: original object reference; \textbf{Mod. Ref.}: reference after modulation by the reference modulator; \textbf{Track. Traj.}: executed object trajectory.}
  \vspace{-5pt}
  \label{fig:Modulation-reference}
\end{figure}

\begin{figure}[!htbp]
  \centering
  \includegraphics[width=0.9\linewidth]{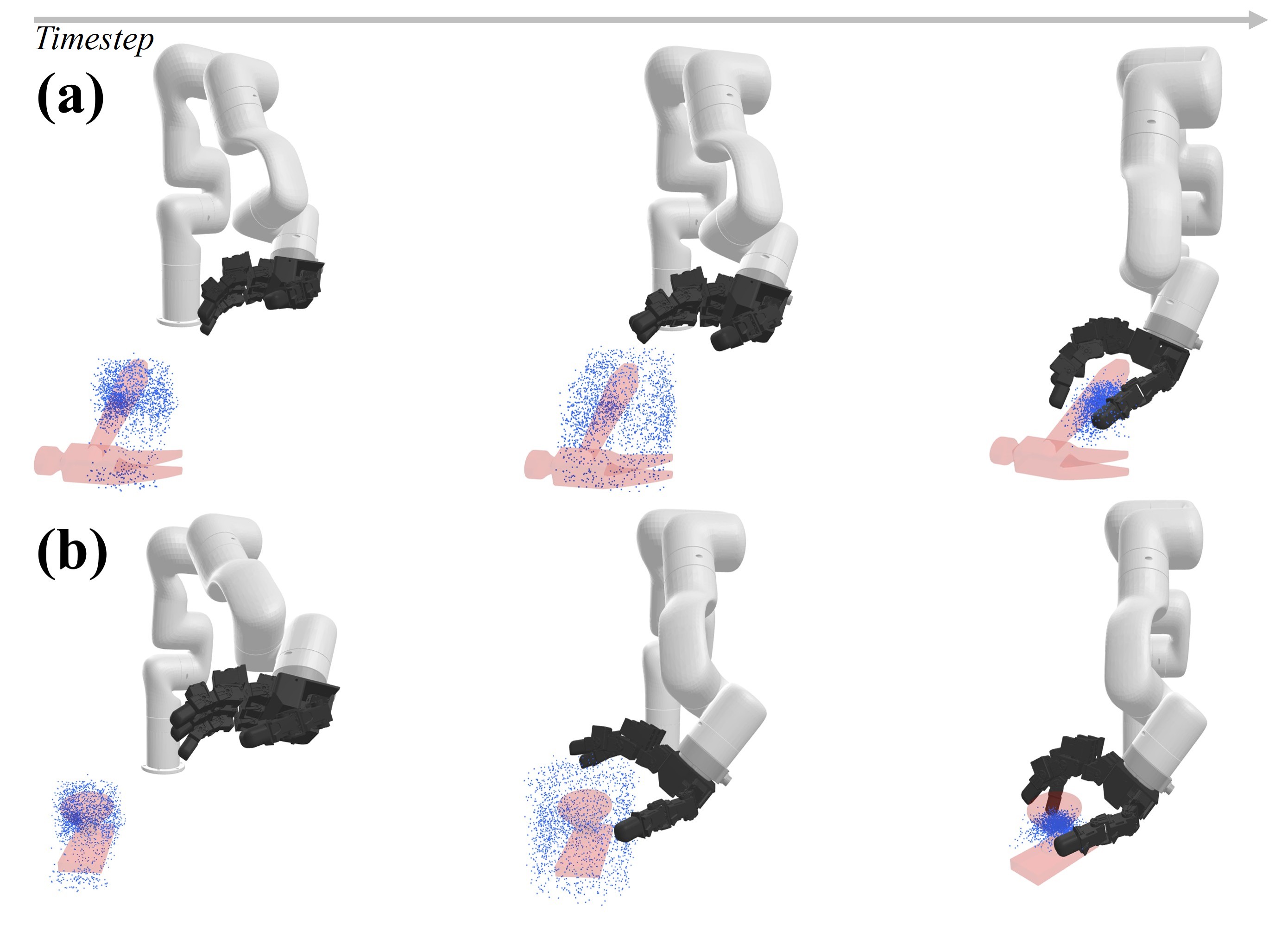}
  \vspace{-10pt}
    \caption{Object latent modulation (General Tracker+O). \textcolor{red}{Red}: actual object and its pose. \textcolor{blue}{Blue}: point cloud decoded from the modulated object latent and its pose. (a) Hammer; (b) Stamp.}

  \vspace{-8pt}
  \label{fig:Modulation-object_latent}
\end{figure}

To further demonstrate the effectiveness of each component, we present qualitative results. In the reference modulation (General Tracker+R) setting, Fig.~\ref{fig:Modulation-reference} shows that when the tracker drifts, the reference modulator detects the deviation and updates the reference; a clear signature is the updated reference intersecting the executed path in the x–y plane, creating a low-error rejoin point that pulls the policy back onto a feasible trajectory and stabilizes long-horizon tracking. This illustrates the power of reference modulation: by continuously adjusting the tracker’s reference, it prevents the error accumulation inherent to a fixed reference and delivers more robust tracking. In the object latent modulation (General Tracker+O) setting, Fig.~\ref{fig:Modulation-object_latent} shows a simple yet strong effect of the object latent modulation: the object latent acts like a ``knob'' for hand openness—a larger latent makes the hand more open, a smaller latent makes it tighter. The modulator uses this ``knob'' phase by phase: during reach, it prefers larger latents to keep the fingers open and avoid bumping the object; during grasp, it switches to smaller latents to close the hand and secure contact. This phase-aware selection shows that the object latent is a meaningful, semantically aligned handle for invoking and composing skills from the tracker’s library, enabling safer approach and more stable grasping—and improving tracking—without changing the reference or issuing large action corrections.

\begin{figure}[!htbp]
  \centering
  \vspace{-5pt}
  \includegraphics[width=0.39\textwidth]{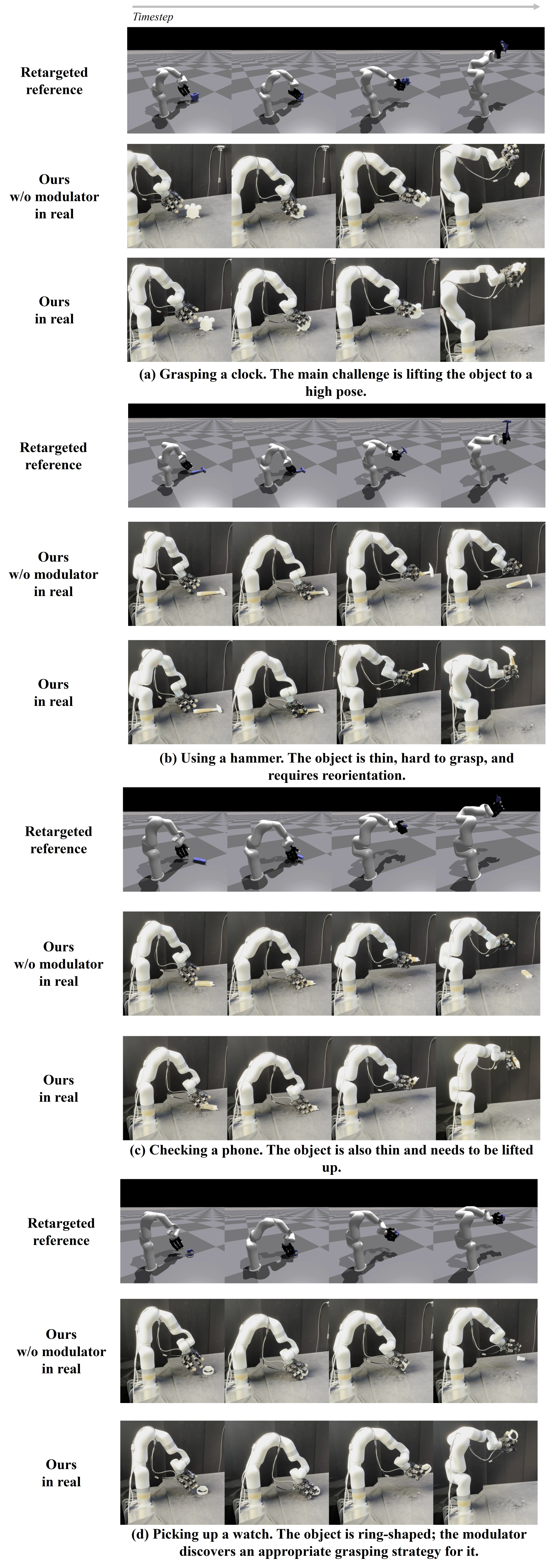}
  \vspace{-5pt}
  \caption{\textbf{Qualitative results.} Panels (a–d): given a noisy text-generated reference retargeted to the robot, \textbf{AdaDexTrack} successfully tracks the long-horizon interaction, while the tracker-only baseline fails. This is because the in-loop modulator continuously adapts the reference, object latent, and positional target, mitigating sim-to-real noise and preventing error accumulation, whereas the tracker-only baseline accumulates errors over time.}

  \vspace{-12pt}
  \label{fig_qualitive_results}
\end{figure}

\noindent\textbf{Data-Scaling Analysis.}
To study the effect of data scale, we vary only the training set size by downsampling the full dataset (1.00×) to a range of log-spaced fractions and train a separate model from scratch for each scale (each scale is the fraction of the original set used). We evaluate on the unseen-trajectory split (Fig.~\ref{fig:scaling1}) and the unseen-object split (Fig.~\ref{fig:scaling2}). In both cases, performance increases steadily with data size and surpasses the tracker-only baseline.

\subsection{Sim-to-Real Experiments}
\label{sec:real}

\noindent\textbf{Sim-to-Real Setup.}
We deploy the policy trained entirely in simulation for closed-loop control in the real world (zero-shot). The hardware mirrors the simulation: an XArm6~\cite{UFactory_xArm_UM_v2_0_0_2023} paired with a LEAP Hand~\cite{shaw2023leap}. Object pose is estimated by FoundationPose~\cite{wen2024foundationpose}, and hand–eye calibration is performed via the Tsai--Lenz method~\cite{tsai1989new}. Because FoundationPose exhibits rotational ambiguity on axially symmetric objects (e.g., bottles, where the estimate can freely spin around the symmetry axis), we exclude those objects to avoid confounding perception errors. We evaluate on two real-world sets: (i) an unseen-trajectory set with 60 reference trajectories from the unseen-trajectory split, covering 15 objects; and (ii) an unseen-object set with 20 reference trajectories from the unseen-object split, covering 2 objects.
For real-world evaluation, we report the completion rate, i.e., the Completion variant of Success Rate in Sec.~\ref{sec:setup}, using identical thresholds and averaged over trajectories.

\begin{figure}[!htbp]
  \centering
  \vspace{-5pt}
  \begin{subfigure}{0.49\linewidth}
    \includegraphics[width=\linewidth]{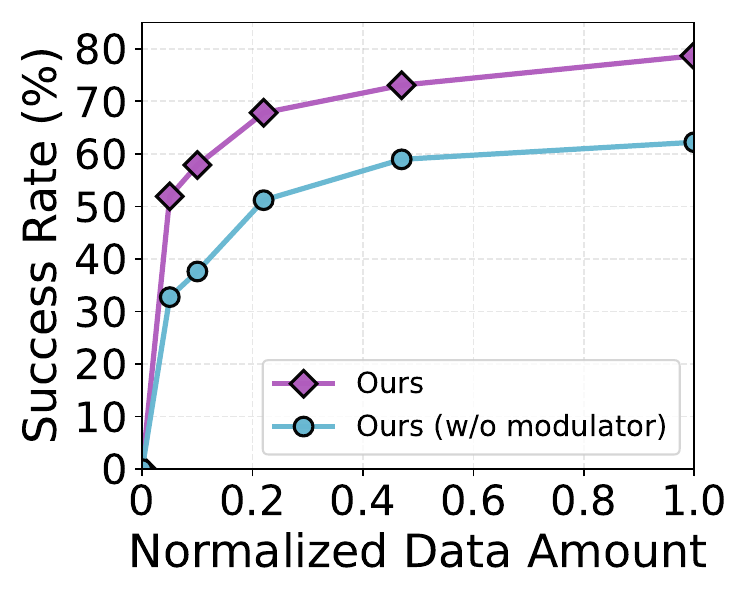}
    \caption{Unseen Trjectories.}
    \label{fig:scaling1}
  \end{subfigure}
  \hfill
  \begin{subfigure}{0.49\linewidth}
    \includegraphics[width=\linewidth]{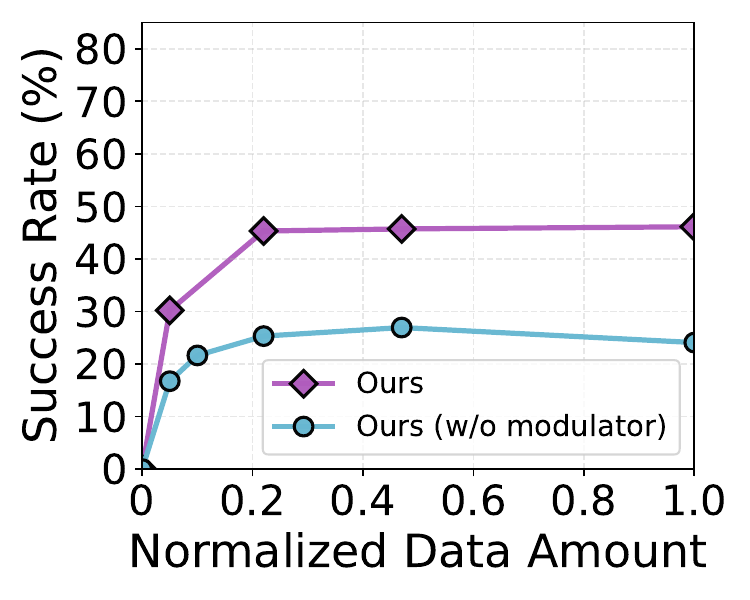}
    \caption{Unseen Objects.}
    \label{fig:scaling2}
  \end{subfigure}
  \vspace{-5pt}
  \caption{Effect of scaling the amount of reference data.}
  \vspace{-15pt}
  \label{fig:scaling-comparison}
\end{figure}

\noindent\textbf{Results.}
As reported in Table~\ref{tab:sim2real}, our approach outperforms the tracker-only baseline in zero-shot real-world tests on both the unseen-trajectory and unseen-object sets. Qualitative results in Figs.~\ref{fig:teaserrrrrrrr} and \ref{fig_qualitive_results} illustrate the benefits of the modulator compared with the tracker-only baseline. The gains come from the in-the-loop modulator, which performs reference modulation, object latent modulation, and small, bounded positional target corrections—counteracting perception noise, calibration drift, latency, and other sim-to-real mismatches before they accumulate. By contrast, the baseline keeps the reference outside the loop and lacks corrections at the reference, representation, and actuation levels, so errors compound over long horizons and lead to failures. Qualitative results in Fig.~\ref{fig:scaling-comparison} show more stable contacts, smoother transitions, and better recovery from disturbances.

\begin{table}[t]
\centering
\small
\setlength{\tabcolsep}{8pt} 
\setlength{\abovecaptionskip}{2pt}   
\begin{tabular}{@{}ccc@{}}
\toprule
\textbf{Methods} & \textbf{Unseen Traj} & \textbf{Unseen Obj} \\
\midrule
Ours (w/o $\text{modulator}$) & 26.63\% & 22.27\% \\
\textbf{Ours}            & \textbf{52.36\%} & \textbf{36.71\%} \\
\bottomrule
\end{tabular}
\caption{Zero-shot sim-to-real experiments. Completion rate (\%).}
\vspace{-15pt}
\label{tab:sim2real}
\end{table}

\section{Conclusions and Limitations}
We introduced \textbf{AdaDexTrack}, a modulator-in-the-loop framework for language-guided HOI tracking. The tracker serves as the skill carrier distilled from many experts, while the modulator provides three tightly coupled interfaces—reference modulation, object-latent modulation, and positional target modulation. Trained under the same task objective as the tracker, these interfaces compose skills on demand, close the loop around the reference, and suppress long-horizon drift. \textbf{AdaDexTrack} surpasses baselines on unseen trajectories and unseen objects, improves with dataset scale, and achieves zero-shot sim-to-real transfer.

\noindent\textbf{Limitations.} In the real-world setting, our system relies on an explicit 6-DoF pose estimator and thus struggles with symmetric objects, where pose is ambiguous. A pose-free route is to drive tracking and modulation directly from point clouds by learning symmetry-aware, SE(3)-equivariant embeddings for reference matching and latent modulation. 

 \clearpage
{
    \small
    \bibliographystyle{ieeenat_fullname}
    \bibliography{main}
}

\clearpage
\setcounter{page}{1}
\appendix  
\maketitlesupplementary




\refstepcounter{part}%
\addcontentsline{toc}{part}{Supplementary Material}%

\etocsetnexttocdepth{subsection}           
\etocsettocstyle{\section*{Contents}}{}    
\localtableofcontents

\section{Additional Details on Method}


\subsection{Data Preprocessing}
\begin{figure}[!htbp]
  \centering
  \vspace{-5pt}
  \begin{subfigure}{0.48\linewidth}
    \centering
    \includegraphics[width=\linewidth]{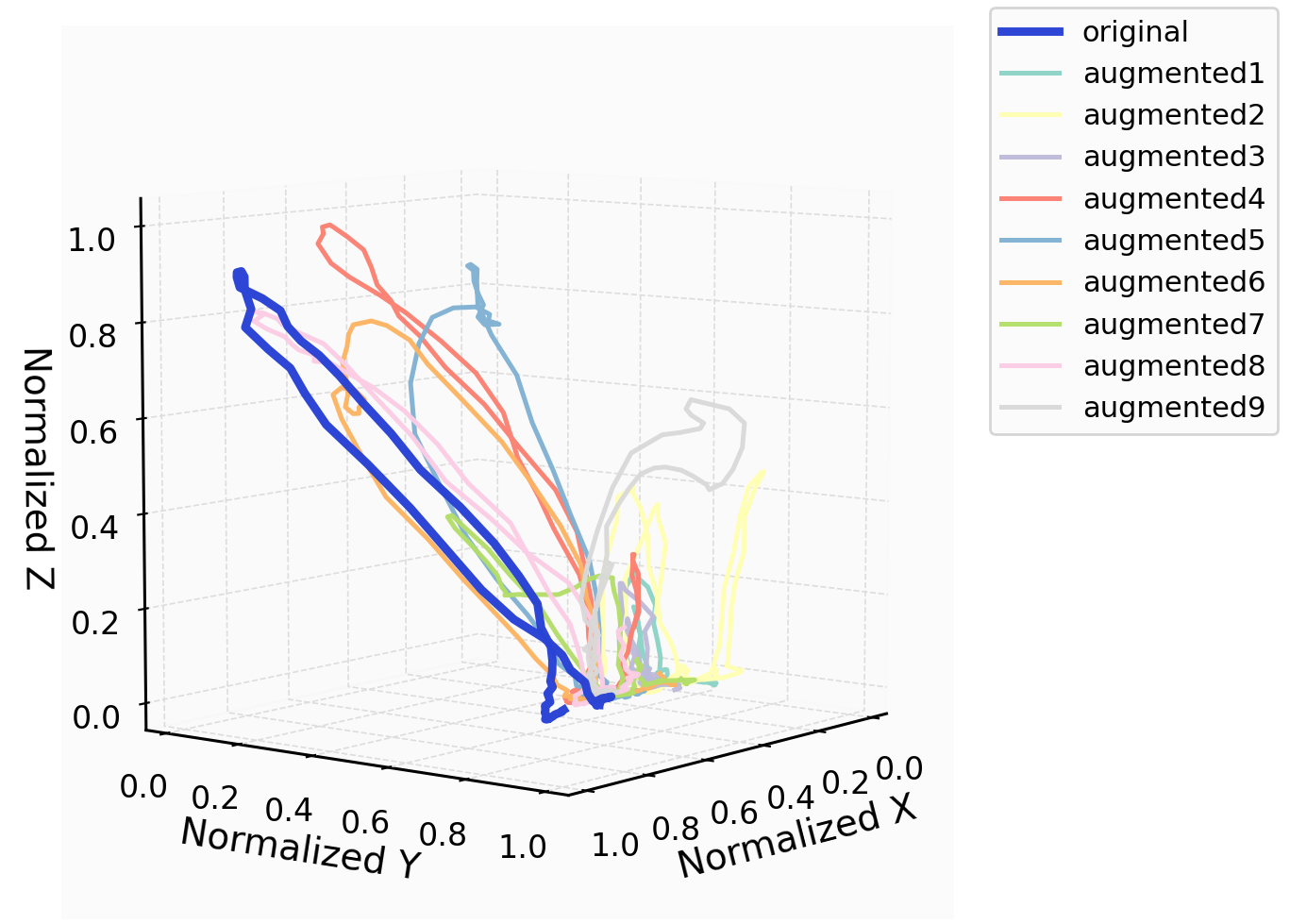}
    \caption{3D view}
    \label{fig:data_aug_3d}
  \end{subfigure}
  \hfill
  \begin{subfigure}{0.48\linewidth}
    \centering
    \includegraphics[width=\linewidth]{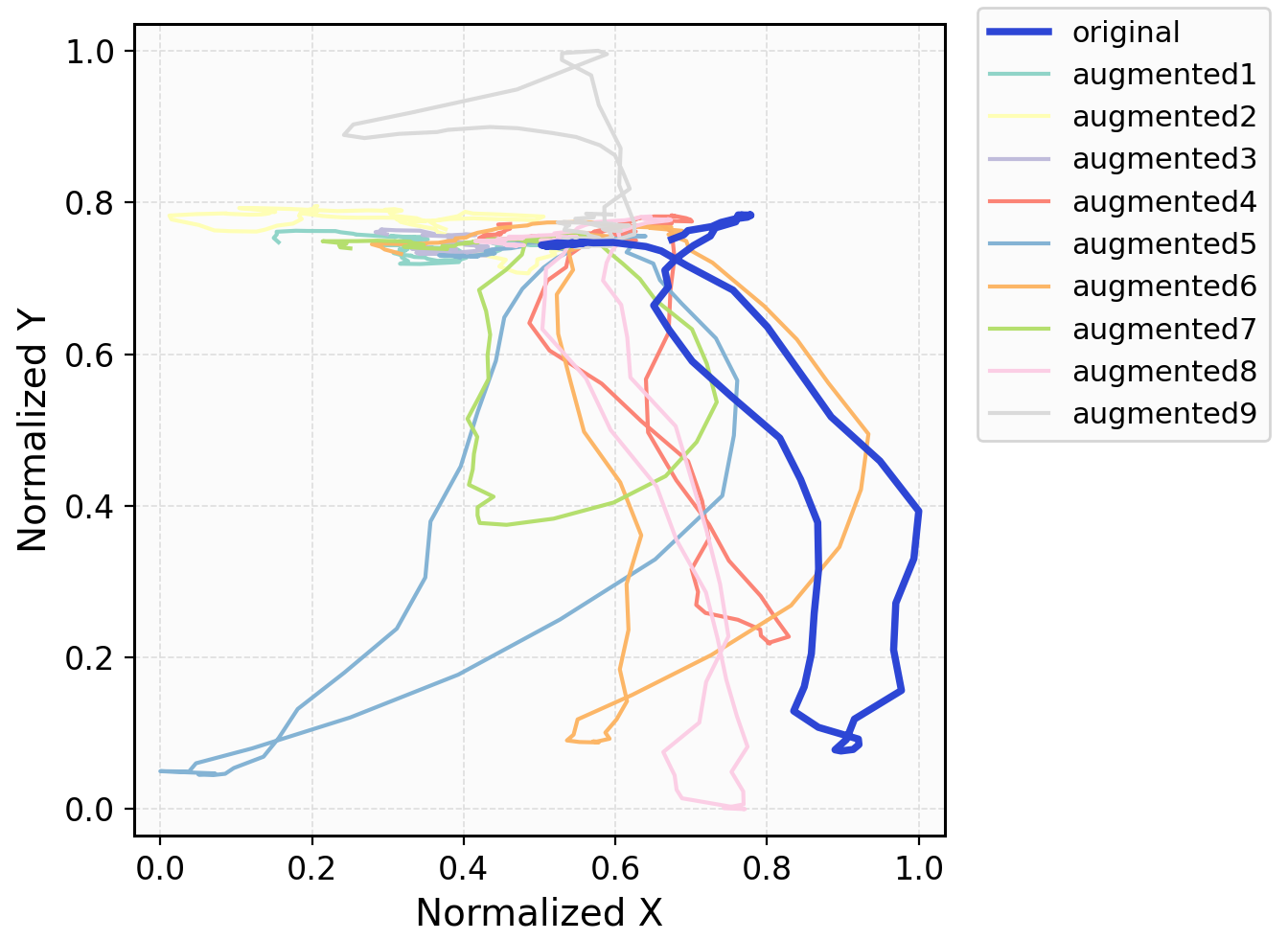}
    \caption{Top-down XY view}
    \label{fig:data_aug_xy}
  \end{subfigure}
  \vspace{-5pt}
  \caption{Language-guided HOI reference augmentation in 3D and top-down XY views.}
  \vspace{-2pt}
  \label{fig:data_aug}
\end{figure}

\noindent\textbf{Language-Guided Trajectory Augmentation.}
We build our language-guided HOI trajectory dataset on top of the text-annotated sequences provided by DiffH2O~\cite{christen2024diffh2o}. Starting from the full set of demonstrations, we retain only sequences that (i) involve single-hand (predominantly right-hand) manipulation without bimanual interaction, (ii) exhibit no hand–robot-arm collisions, and (iii) keep the hand pose within the manipulator’s reachable workspace at every timestep, which yields 505 reference trajectories. For each retained trajectory, we then perform paraphrase-based language augmentation. Concretely, we use GPT-5~\cite{openai_chatgpt5_2025} to generate nine semantically equivalent descriptions that follow DiffH2O’s native annotation format and preserve the original task intent. We feed these paraphrased descriptions back into the DiffH2O text-to-motion model to synthesize additional HOI trajectories conditioned on the same object geometry. Fig.~\ref{fig:data_aug} shows a qualitative example of our augmentation results, where a single original trajectory is expanded into ten variants. Applying the same feasibility filters to the augmented set results in 2,765 valid text–trajectory pairs spanning 50 objects.


\noindent\textbf{Kinematic Retargeting.}
Our kinematic retargeting procedure is similar to DexTrack~\cite{liu2025dextrack}. 
Given a human hand–object interaction trajectory with MANO hand poses $\{\mathbf{H}^{\text{human}}_n\}$ and object poses $\{\mathbf{O}_n\}$, we retarget it to the robot hand by matching keypoints between the MANO hand and the articulated robot hand. 
Let $\mathbf{K}^{\text{human}}_n$ be the MANO keypoints at timestep $n$ and $\mathbf{K}_n(\boldsymbol{\theta}_n)$ the corresponding robot hand keypoints obtained via forward kinematics from joint angles $\boldsymbol{\theta}_n$. 
At each timestep, we solve a least-squares problem $\min_{\boldsymbol{\theta}_n} \|\mathbf{K}_n(\boldsymbol{\theta}_n) - \mathbf{K}^{\text{human}}_n\|$, yielding a sequence of robot hand joint poses $\{\boldsymbol{\theta}_n\}$ that serves as the kinematic reference for our controller.

\noindent\textbf{Object Latent Encoding.}
We encode object geometry with a latent code extracted by a Chamfer-trained point-cloud autoencoder. Given an object point cloud $P$, the encoder produces the object latent $f_o = E_{pc}(P)$, which is used as the object representation in our tracker and modulated online during execution.

\subsection{Sim-to-Real Transfer}
\begin{figure}[!htbp]
  \centering
  \vspace{-5pt}
  \includegraphics[width=1\linewidth]{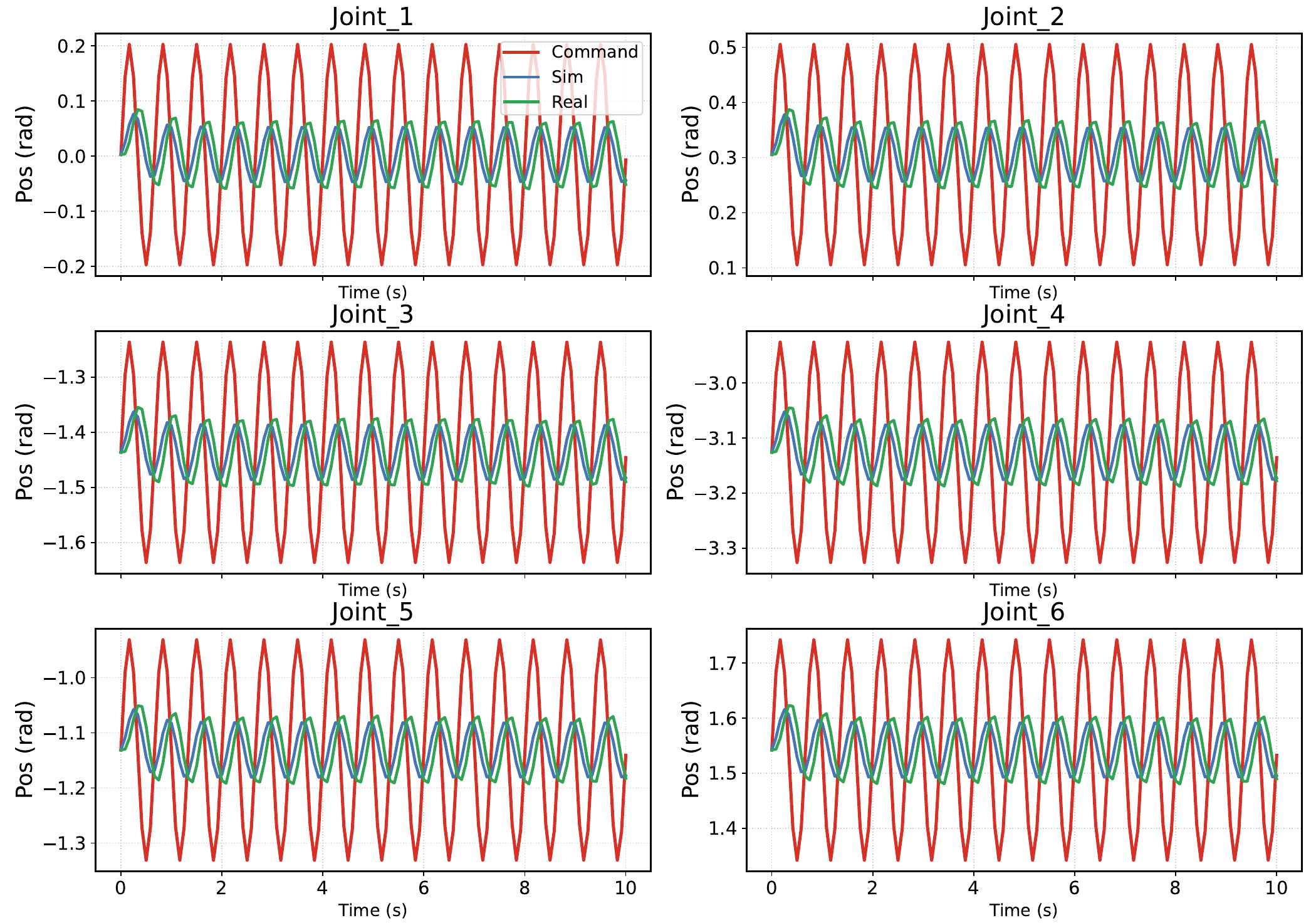}
  \vspace{-15pt}
\caption{System identification of the arm joints. For each joint, we plot three traces: (1) the commanded input, (2) the simulated response using the identified dynamics model, and (3) the measured real response. The close agreement between (2) and (3) indicates that the identified parameters faithfully capture the joint dynamics.}
    \vspace{10pt}
  \label{fig:sys_id}
\end{figure}

\noindent\textbf{System Identification.}
We perform system identification on the robot arm and hand by driving each joint with a family of sinusoidal position commands and recording the resulting joint trajectories. 
Specifically, we apply commands of the form $A(t) = a \sin(2\pi f t)$, where both the frequency $f$ and the amplitude $a$ are swept over a range, and $a$ is chosen such that the commanded motion stays within the joint limits. 
We then replay the same command sequences in simulation and identify, for each joint independently, the stiffness and damping parameters by minimizing the discrepancy between the simulated and real joint trajectories. 
As shown in Fig.~\ref{fig:sys_id}, the identified model produces response curves in simulation that closely match those of the real robot under identical control commands.

\noindent\textbf{Domain Randomization.}
To improve robustness and facilitate sim-to-real transfer, we apply domain randomization to both the system dynamics and the observations/actions during training. 
On the dynamics side, we randomly perturb the arm and hand parameters, including link masses, joint friction, joint stiffness, and joint damping, around their nominal values. 
For the manipulated objects, we use a larger range of randomization over physical parameters such as mass, friction coefficients, and restitution, and also randomize the initial object pose relative to the reference trajectory. 
On the sensing and control side, we inject Gaussian noise into the proprioceptive states and add small Gaussian noise to the action commands. 
The exact ranges used for domain randomization are summarized in Table~\ref{tab:domain_rand}.


\begin{table}[t]
\vspace{-5pt}
\centering
\small              
\setlength{\tabcolsep}{4pt} 
\begin{tabular}{llll}
\toprule
Parameter & Type & Distribution & Initial Range \\
\midrule
\multicolumn{4}{l}{\textbf{Robot}} \\
Mass              & Scaling  & uniform     & $[0.9, 1.1]$   \\
Friction          & Scaling  & uniform     & $[0.9, 1.1]$   \\
Joint stiffness   & Scaling  & uniform     & $[0.9, 1.1]$   \\
Joint damping     & Scaling  & uniform     & $[0.9, 1.1]$   \\
\midrule
\multicolumn{4}{l}{\textbf{Object}} \\
Mass                     & Scaling  & uniform  & $[0.7, 1.3]$  \\
Friction                 & Scaling  & uniform  & $[0.7, 1.3]$  \\
Restitution              & additive  & uniform  & $[0.0, 0.4]$  \\
Scale                    & Scaling  & uniform  & $[0.95, 1.05]$ \\
Initial Position (x,y)   & Additive & Gaussian & $[0.0, 0.05]$ \\
Initial Rotation (yaw)   & Additive & Gaussian & $[0.0, 0.5]$  \\
\midrule
\multicolumn{4}{l}{\textbf{Observation}} \\
Proprioception           & Additive & Gaussian & $[0.0, 0.002]$ \\
\midrule
\multicolumn{4}{l}{\textbf{Action}} \\
Correlated noise         & Additive & Gaussian & $[0.0, 0.02]$ \\
Uncorrelated noise       & Additive & Gaussian & $[0.0, 0.05]$ \\
\bottomrule
\end{tabular}
\caption{Domain randomization configuration.}
\label{tab:domain_rand}
\vspace{-10pt}
\end{table}

\subsection{Tracking Controller Training}
\subsubsection{Specialist Policy}
\noindent\textbf{Action Space.}
The control action at each timestep is defined in joint space. 
We use a 22-DoF arm–hand system with joint configuration $\mathbf{q} \in \mathbb{R}^{22}$. 
The action space $\mathcal{A} \subset \mathbb{R}^{22}$ consists of relative joint commands
$\Delta \mathbf{q} \in \mathbb{R}^{22}$. 
Internally, we maintain a vector of joint position targets $\mathbf{q}^{\text{tar}}$. 
At every control step, the policy outputs $\Delta \mathbf{q}$ and we update the targets by
$\mathbf{q}^{\text{tar}} \leftarrow \mathbf{q}^{\text{tar}} + \Delta \mathbf{q}$ (followed by clipping to the joint limits). 
The updated targets $\mathbf{q}^{\text{tar}}$ are then sent to the low-level joint position controllers as setpoints.

\noindent\textbf{Observation Space.} Table~\ref{tab:spec_obs} details the observation space used for training the specialist policies.

\noindent\textbf{Reward implementation details.}
In practice, we implement the reward in Eq.~\eqref{eq:rt-loss} as a shaped combination of tracking and proximity terms. 
The joint and hand-pose terms are realized by a tracking loss $\ell_{\text{track}}(t)$ that aggregates palm position and orientation errors, 22-DoF joint-space errors, and fingertip position errors with respect to their references; this loss enters the reward with a negative weight so that larger deviations from the reference are penalized more strongly. 
The hand--object proximity term is instantiated using the distances between the palm and the object, and between the fingertips and the object, which encourages the hand to move into and remain in the vicinity of the object. 
To avoid excessively large gradients, these distances are clipped to fixed upper bounds, and the fingertip penalty is activated only once the palm is sufficiently close to the object.

The object-pose term in Eq.~\eqref{eq:rt-loss} is implemented via a goal distance $d_{\text{goal}}(t)$ that combines translational and rotational discrepancies between the current and target object poses. 
We represent rotations as unit quaternions and compute the angular component of $d_{\text{goal}}(t)$ from the quaternion difference via the dot product between the normalized goal quaternion and the current quaternion. 
In addition, we include a sparse success bonus 
$r_{\text{bonus}}(t) = \mathbb{I}\bigl[d_{\text{palm}}(t)\!\leq\!\tau_{\text{palm}},\, d_{\text{fing}}(t)\!\leq\!\tau_{\text{fing}},\, d_{\text{goal}}(t)\!\leq\!\tau_{\text{goal}}\bigr]\,(1 + 10\, d_{\text{goal}}(t))^{-1}$, 
where $\mathbb{I}[\cdot]$ is the indicator function. 
This bonus rewards states where the hand remains close to the object and the object pose is very close to the goal, and can be viewed as a shaped realization of the proximity component in Eq.~\eqref{eq:rt-loss}.

To improve computational efficiency and obtain a smooth orientation error, we represent all orientations using unit quaternions and compute $d_R(\cdot)$ from the quaternion difference via the dot product between the normalized goal quaternion and the current quaternion.

\noindent\textbf{Semantic Grouping of Specialists.}
We start from the text-annotated DiffH2O demonstrations and first apply a feasibility filter (single-hand interaction, no collisions, reachable workspace), which leaves 505 valid human--object interaction sequences. 
On top of these 505 base demonstrations, we then perform paraphrase-based augmentation: for each retained sequence, we use GPT-5 to generate multiple semantically equivalent but textually different captions and feed them into DiffH2O to synthesize additional reference trajectories. 
This procedure yields at most ten references per base demonstration (one from the original caption and up to nine from paraphrased captions), i.e., 5050 candidates in total. 
We apply the same feasibility filter again and obtain 2{,}765 valid reference trajectories overall.

For specialist training, we group references by their underlying semantic intent: all trajectories that originate from the same base demonstration (and its paraphrased captions) are treated as one semantic group. 
Each such group defines a single specialist tracker $\pi_i^{\text{track}}$, which is trained jointly on all of its associated reference trajectories by running them in parallel in Isaac Gym.
In this way, we obtain one specialist per semantic group, with each specialist responsible for a small cluster of semantically equivalent references rather than a single trajectory. 
This semantic grouping substantially reduces the number of specialist policies while still exposing each of them to multiple motion and language variants of the same underlying task.


\noindent\textbf{Time consumption.} Each specialist is trained in Isaac Gym~\cite{makoviychuk2021isaac} with 8{,}192 parallel environments on a single NVIDIA A10 GPU and converges in about 3 hours.

\subsubsection{Generalist Policy}
The observation space used for behavior cloning of the generalist policy is identical to that of the specialist policy, as summarized in Table~\ref{tab:spec_obs}.

\noindent\textbf{Time consumption.} The generalist policy is trained on eight NVIDIA A10 GPUs and converges in about two days.

\begin{table}[t]
\centering
\begin{tabular}{c|p{0.65\linewidth}}
\hline
Index & Description \\
\hline
0 -- 22   & arm-hand joint positions \\
\hline
22 -- 57 & the palm pose and the poses of the four fingertips  \\
\hline
57 -- 64 & object pose \\
\hline
64 -- 71 & distance between the current object pose and the object reference pose at the current timestep \\
\hline
71 -- 99 & object reference poses at future timesteps $\{1,2,4,12\}$ relative to the current timestep \\
\hline
99 -- 121 & distance between the current and reference arm–hand joint positions at the current timestep \\
\hline
121 -- 209 & reference arm–hand joint positions at future timesteps $\{1,2,4,12\}$ relative to the current timestep \\
\hline
209 -- 273 & object latent \\
\hline
\end{tabular}
\caption{Observation space used for specialist policy training.}
\label{tab:spec_obs}
\vspace{-10pt}
\end{table}



\subsection{Tracking Modulator Training.}
During modulator training, we freeze the generalist policy and optimize only the modulator. 
The modulator uses the same observation space as in Table~\ref{tab:spec_obs}, and its reward is identical to those used in specialist training. 
As discussed in Sec.\ref{sec:modulator}, the modulator modifies both the inputs and outputs of the tracker: it modulates the tracking reference and the object latent in the observation, and applies a small residual modulation to the tracker’s action.

\noindent\textbf{Time consumption.} The modulator policy is trained in Isaac Gym~\cite{makoviychuk2021isaac} with 8{,}192 parallel environments on a single NVIDIA A10 GPU and converges in about two days.

\begin{table}[t]
\vspace{5pt}
\centering
\begin{tabular}{c|p{0.65\linewidth}}
\hline
Index & Description \\
\hline
0 -- 22   & arm-hand joint positions \\
\hline
22 -- 57 & the palm pose and the poses of the four fingertips  \\
\hline
57 -- 64 & object pose \\
\hline
64 -- 71 & distance between the current object pose and the object reference pose at the current timestep \\
\hline
71 -- 141 & object reference poses at future ten timesteps relative to the current timestep \\
\hline
141 -- 147 & distance between the current and reference arm joint positions at the current timestep \\
\hline
147 -- 207 & reference arm joint positions at future ten timesteps relative to the current timestep \\
\hline
207 -- 271 & object latent \\
\hline
\end{tabular}
\caption{Observation space for the low-level controller in ObjDex.}
\label{tab:spec_obs_objdex}
\end{table}

\section{Simulation Experiments Details}
\subsection{Environment Setup}
We use the Isaac Gym simulator with a simulation frequency of 60\,Hz and a control frequency of 12\,Hz. 
For each reference trajectory to be tracked, we define the tracking start frame as the first frame where the minimum distance between the hand mesh and the object center exceeds 20\,cm. 
Starting from this frame, we take the subsequent 300 frames as the full trajectory segment used for training.

\subsection{Baseline Implementation}
\subsubsection{DexTrack}
We implement a single, generalist neural tracker that takes the current robot/object state and a fixed reference as input and directly outputs actions—without any in-loop modulation (no reference, object latent, or positional target modulation at test time). To ensure a fair comparison, we keep the observation/action spaces, reference format, rewards, training data, and evaluation protocol identical to those used for AdaDexTrack. The object latent is computed once at episode start and held fixed throughout execution, matching the fixed-reference, fixed-representation setting.

\noindent\textbf{Imitation targets (teacher actions).}
To supply the IL term, we reuse the specialist tracking policies trained in Sec.\ref{sec:tracking}. For each training reference, we roll out its matched specialist in simulation and log synchronized tuples
\((\mathbf{o}_t,\ \hat{\mathbf{s}}_t,\ \mathbf{a}^{\text{teacher}}_t)\),
where \(\mathbf{o}_t\) is the policy observation, \(\hat{\mathbf{s}}_t\) is the time-indexed HOI reference, and \(\mathbf{a}^{\text{teacher}}_t\) is the specialist’s action at time \(t\). These teacher actions form the imitation dataset used by DexTrack’s IL term.

\noindent\textbf{Training objective (IL+RL).}
The combined IL+RL training procedure and the homotopy-based demonstration mining described in DexTrack are not publicly available. We therefore re-implement the IL+RL scheme (without homotopy mining): we train the policy with PPO on the task reward and add a time-indexed imitation regularizer that directly matches the policy's action at timestep $t$ to the logged teacher action $\mathbf{a}^{\text{teacher}}_t$ for that same reference index.
In other words, since tracking uses a time-indexed reference, the imitation target is keyed by $t$ (no observation-based matching).
The total objective is the standard PPO actor--critic loss plus a weighted imitation term.

\subsubsection{ObjDex}
We reproduce a hierarchical planner-over-controller baseline: a high-level, object-centric planner synthesizes wrist trajectories conditioned on the task goal, and a low-level RL controller tracks these trajectories with the dexterous hand. The two layers are trained with different objectives (generative trajectory modeling vs.\ RL) and communicate primarily via wrist-space references.


\noindent\textbf{High-level planner.}
Since the original planner code is unavailable, we re-implement it as a Transformer-based conditional trajectory generator.
The planner takes the object ID and the object reference trajectory as input and outputs a wrist reference trajectory for the low-level controller to track.
We train it in a supervised fashion on pairs of object IDs and object reference trajectories, using the wrist reference trajectory as the target.
At test time, given an object ID and its reference trajectory, the planner predicts the wrist reference.
For unseen objects, following ObjDex, we condition the planner on a seen surrogate object ID from the same category and predict the wrist reference trajectory for the unseen object.

\noindent\textbf{Low-level controller.}
We implement an RL tracker with the same action space as AdaDexTrack for fairness, but with a different observation design. Following ObjDex, the policy receives a 10-step lookahead reference containing only the object pose and arm joint positions for the next ten timesteps (see Table~\ref{tab:spec_obs_objdex}). Following ObjDex to mitigate embodiment gap, the reward removes fingertip tracking and hand joint position tracking terms, and retains only (i) wrist pose tracking and (ii) arm joint position tracking. At test time, the controller takes the current state and the planner’s wrist trajectory and tracks it in closed loop; no reference re-planning, object latent adjustment, or positional target correction is applied beyond this stack.

\noindent\textbf{Data augmentation loop.}
Following ObjDex, once the low-level controller is trained, we execute it in simulation under the planner’s guidance, collect successful motion trajectories, add them to the planner’s training set, and fine-tune the high-level planner on the augmented data.

\subsubsection{Vanilla RL (PPO)}
We train a single PPO policy across all training references, using the same action space, observation space, and reward as in the specialist training (Sec.~\ref{sec:tracking}).


\subsection{Qualitative Results}
\begin{figure}[!htbp]
  \centering
  \vspace{-15pt}
  \includegraphics[width=0.8\linewidth]{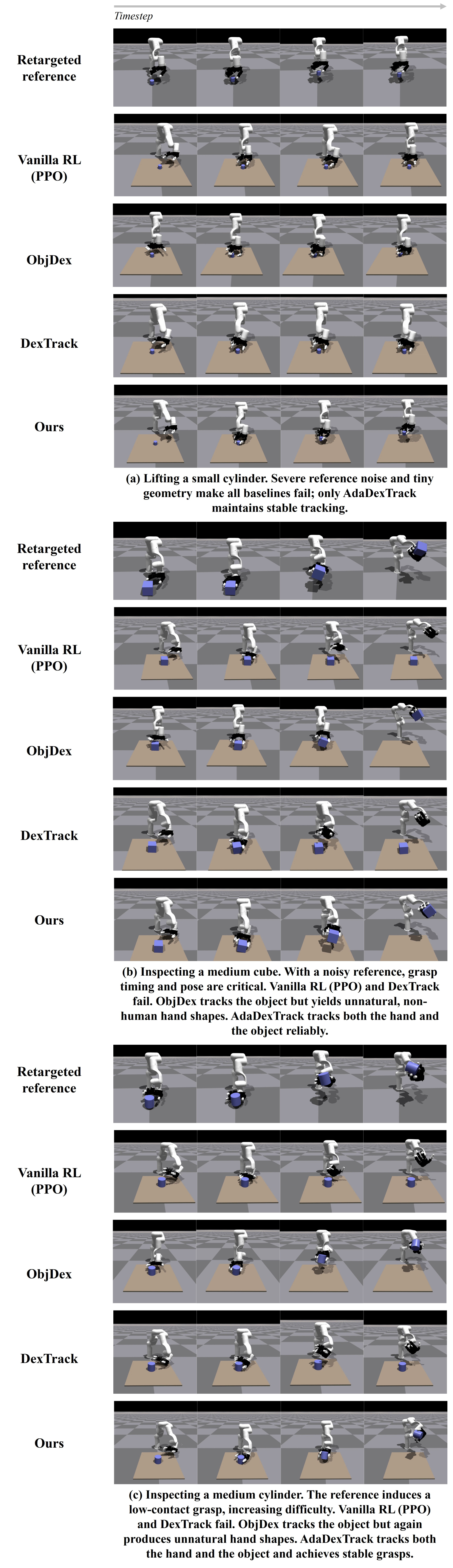}
  \vspace{-8pt}
  \caption{Qualitative results in simulation.}
  \vspace{-2pt}
  \label{fig:sim_qualitative}
\end{figure}

We provide qualitative comparisons with the baselines on the unseen trajectory set in Figure~\ref{fig:sim_qualitative}.
Across tasks, AdaDexTrack achieves the most faithful hand--object tracking, while Vanilla RL (PPO) and DexTrack often drift or fail. ObjDex typically follows the object trajectory but, lacking hand-shape tracking, produces unnatural, non-humanlike hand motions.

\begin{figure}[!htbp]
  \centering
  \vspace{-5pt}
  \includegraphics[width=0.8\linewidth]{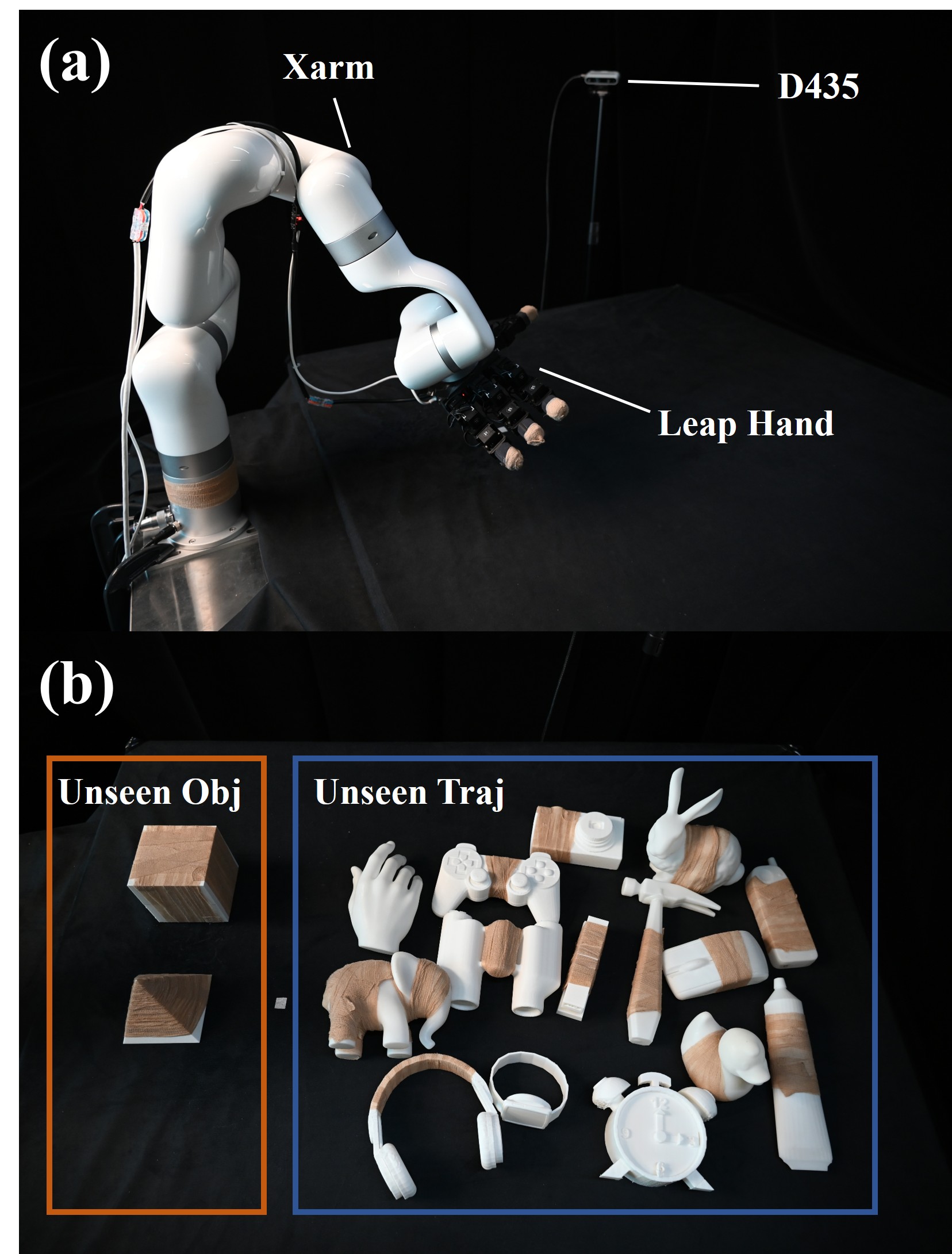}
  \vspace{-5pt}
\caption{Real-world setup. (a) Hardware. (b) Test objects. The blue box marks objects used for the real-world unseen-trajectory set; the orange box marks objects used for the real-world unseen-object set.}
  \vspace{-10pt}
  \label{fig:real_env}
\end{figure}

\begin{figure}[!htbp]
  \centering
  \vspace{-10pt}
  \includegraphics[width=1\linewidth]{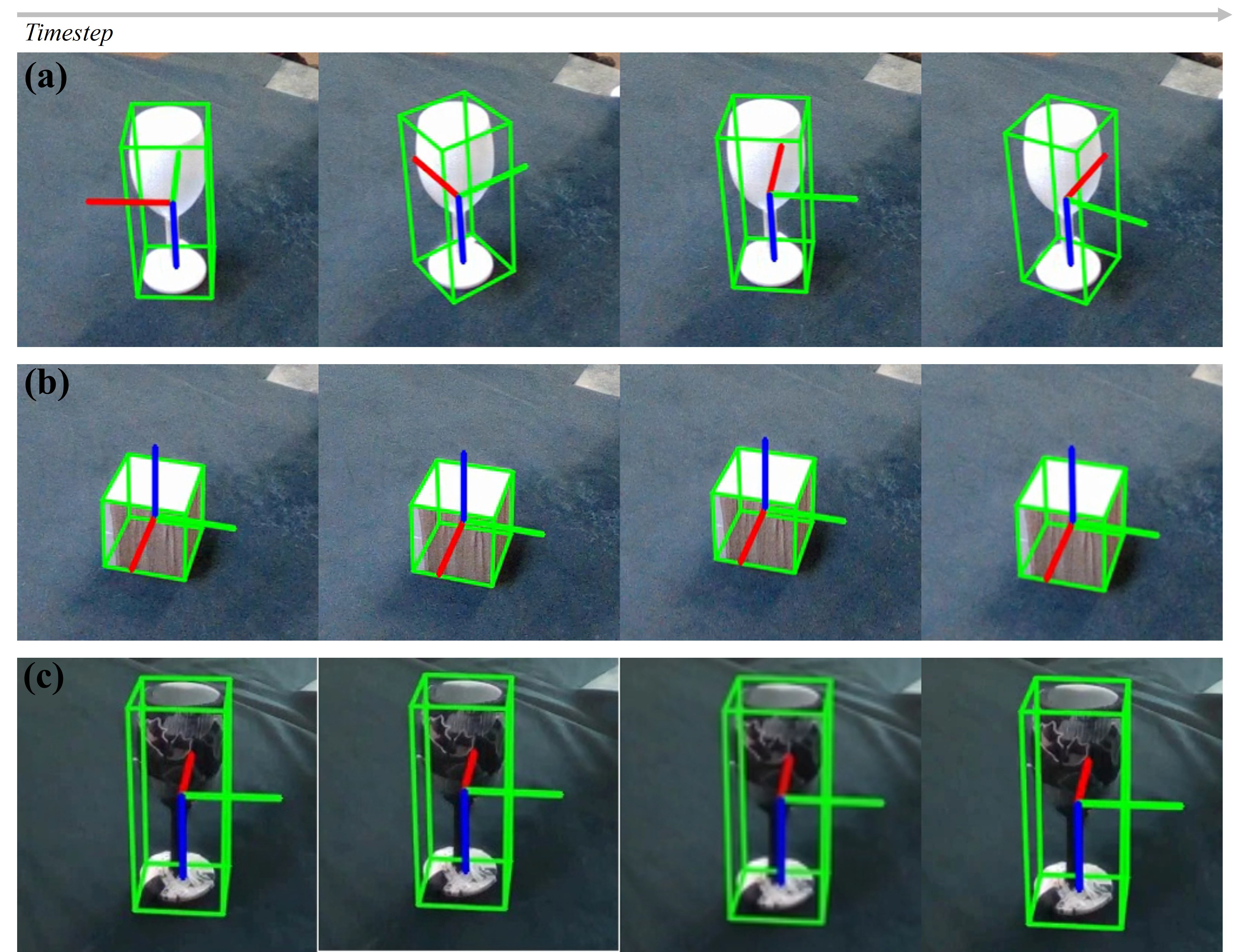}
  \vspace{-10pt}
\caption{FoundationPose estimation behavior. 
(a) Wineglass: although the object is static, the estimate continuously spins around the blue symmetry axis due to axial symmetry. 
(b) Cube: despite discrete symmetry, the pose remains stable (no continuous spin).
(c) Textured wineglass: asymmetric texture breaks the symmetry and yields a stable object frame.}
  \vspace{-2pt}
  \label{fig:founda_failure}
\end{figure}

\section{Real-World Experiments Details}

\subsection{Environment Setup}
Our real-world setup is shown in Fig.~\ref{fig:real_env}. We use an Intel RealSense D435 for RGB-D sensing and FoundationPose~\cite{wen2024foundationpose} for 6-DoF object pose estimation. In the main real-world evaluation, we exclude axially symmetric objects (e.g., wineglass) because FoundationPose exhibits continuous rotational ambiguity on them (Fig.~\ref{fig:founda_failure}a), which would confound the manipulation results. By contrast, objects with discrete symmetries (e.g., cube with $90^\circ$ rotational invariance; Fig.~\ref{fig:founda_failure}b) do not induce continuous spin; edges or mild texture typically yield stable poses up to the object’s symmetry group. Accordingly, we include discretely symmetric objects when the estimator is stable under this equivalence and exclude objects that exhibit free spinning about a symmetry axis. We further study axially symmetric objects with added asymmetric texture in Sec.~\ref{sec:textured_symmetric_objects}, where the texture breaks the visual symmetry and enables a stable object frame (Fig.~\ref{fig:founda_failure}c).

\subsection{Metrics}
In real-world experiments, we report Success Rate as the completion metric defined in Sec.~\ref{sec:setup}, identical to the simulation setting. All thresholds match Sec.~\ref{sec:setup}, and we report the average completion over trajectories.

\subsection{Textured Axially Symmetric Objects}
\begin{figure}[htbp] 
\centering \vspace{-5pt} \includegraphics[width=0.6\linewidth]{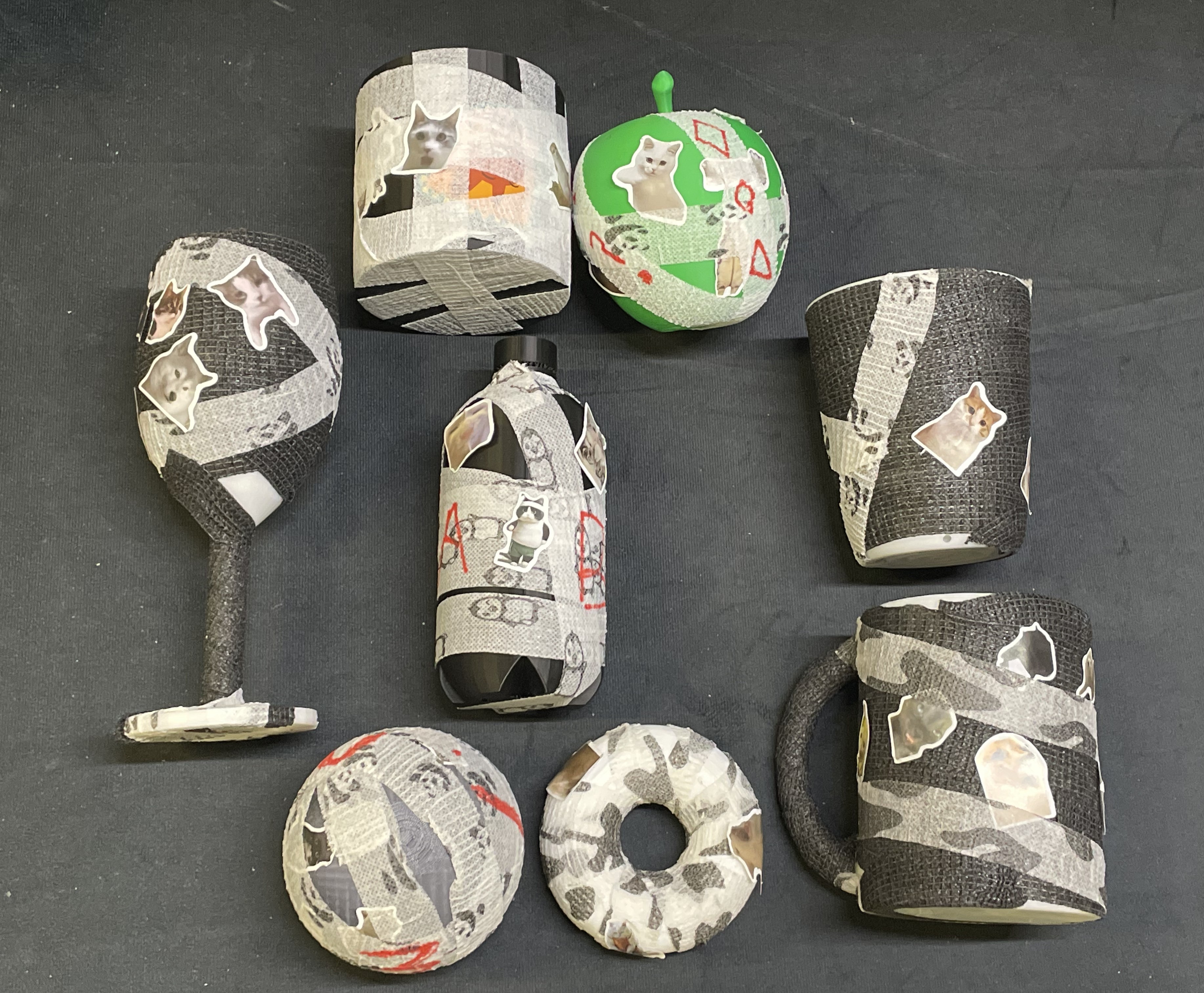} \vspace{-5pt}
\caption{Eight textured axially symmetric objects used in our real-world experiments.} 
\label{fig:textured_symmetric_objects}
\vspace{-10pt} 
\end{figure}

To study axially symmetric objects, we add asymmetric texture to break the rotational symmetry and obtain stable FoundationPose estimates (Fig.~\ref{fig:founda_failure}c). Fig.~\ref{fig:textured_symmetric_objects} shows the 8 textured axially symmetric objects used in this study. Since FoundationPose also requires a known object mesh, we obtain the object meshes using AR Code~\cite{arcode2022}. We evaluate these 8 objects in the real world, with 5 unseen trajectories per object. Table~\ref{tab:textured_symmetric} reports the average completion rate over all trials. With stable pose estimates, our full method substantially outperforms the variant without the modulator, showing that the modulator remains effective on rotationally symmetric geometries.
\label{sec:textured_symmetric_objects}

\begin{table}[htbp]
\vspace{-0.5em}
\centering
\small
\setlength{\tabcolsep}{4pt}
\begin{tabular}{lc}
\hline
\textbf{Method} & \textbf{Completion (\%, $\uparrow$)} \\
\hline
Ours (w/o Modulator) & 29.72\% \\
\textbf{Ours} & \textbf{58.61\%} \\
\hline
\end{tabular}
\caption{Real-world results on textured axially symmetric objects.}
\label{tab:textured_symmetric}
\vspace{-1em}
\end{table}

\subsection{Real-World Evaluation Without Object Tape}
\label{sec:without_object_tape}

To address the concern that tape on the objects may artificially simplify grasping, we conduct an additional real-world evaluation after removing all tape from the objects. We keep the rest of the setup unchanged and re-run both the unseen-trajectory and unseen-object settings. Table~\ref{tab:without_tape} reports the resulting completion rates.

Removing the tape does not degrade performance; instead, it leads to higher success rates for both methods. We attribute this to the fact that excessive friction can hinder the small relative motions needed for grasp adjustment, whereas untaped objects allow more natural micro-sliding during contact. Our full method still substantially outperforms the variant without the modulator, indicating that the improvement does not rely on artificial adhesion, but on more effective handling of natural contact dynamics.

\begin{table}[H]
\vspace{-0.5em}
\centering
\small
\setlength{\tabcolsep}{4pt}
\begin{tabular}{lcc}
\hline
\textbf{Method} & \textbf{Unseen Traj.} & \textbf{Unseen Obj.} \\
\hline
Ours (w/o Modulator) & 36.42\% & 18.75\% \\
\textbf{Ours} & \textbf{65.71\%} & \textbf{32.50\%} \\
\hline
\end{tabular}
\caption{Real-world results without grip tape on the objects. }
\label{tab:without_tape}
\vspace{-1em}
\end{table}

\subsection{Real-World Evaluation with Practical Object Geometry}
\label{sec:practicality_geometry}

To further assess the practicality of our method, we replace the ground-truth object geometry used in the main experiments with inputs obtainable from lightweight perception tools. Specifically, we use InstantMesh~\cite{xu2024instantmesh} to reconstruct object geometry from a single RGB image, and use the resulting geometry for both policy inference and FoundationPose-based pose tracking. We evaluate this setting on the unseen-trajectory and unseen-object sets from Sec.~\ref{sec:real}, without object tape. As shown in Table~\ref{tab:practicality_geometry}, using reconstructed geometry yields performance comparable to ground-truth geometry, indicating that our method does not rely on carefully prepared CAD models.

We further consider a stricter setting where the object input is a segmented partial point cloud from a single-view depth observation. This setting is more challenging due to incomplete geometry and self-occlusion. Nevertheless, our method remains robust, achieving strong performance on both evaluation sets. These results support the practical applicability of our approach under realistic geometry acquisition pipelines.

\begin{table}[H]
\vspace{-0.5em}
\centering
\small
\setlength{\tabcolsep}{4pt}
\begin{tabular}{lccc}
\hline
\textbf{Input} & \textbf{Method} & \textbf{Unseen Traj.} & \textbf{Unseen Obj.} \\
\hline
GT Pointcloud      & Ours & 65.71\% & 32.50\% \\
Recon. Pointcloud  & Ours & 59.28\% & 34.73\% \\
Partial Pointcloud & Ours & 49.46\% & 25.72\% \\
\hline
\end{tabular}
\caption{Real-world results under practical object-geometry inputs. GT Pointcloud uses ground-truth object geometry; Recon. Pointcloud is reconstructed from a single RGB image using InstantMesh~\cite{xu2024instantmesh}; Partial Pointcloud is obtained from segmented single-view depth.}
\label{tab:practicality_geometry}
\vspace{-1em}
\end{table}

\subsection{Qualitative Results}
\begin{figure}[!htbp]
  \centering
  \vspace{-15pt}
  \includegraphics[width=0.8\linewidth]{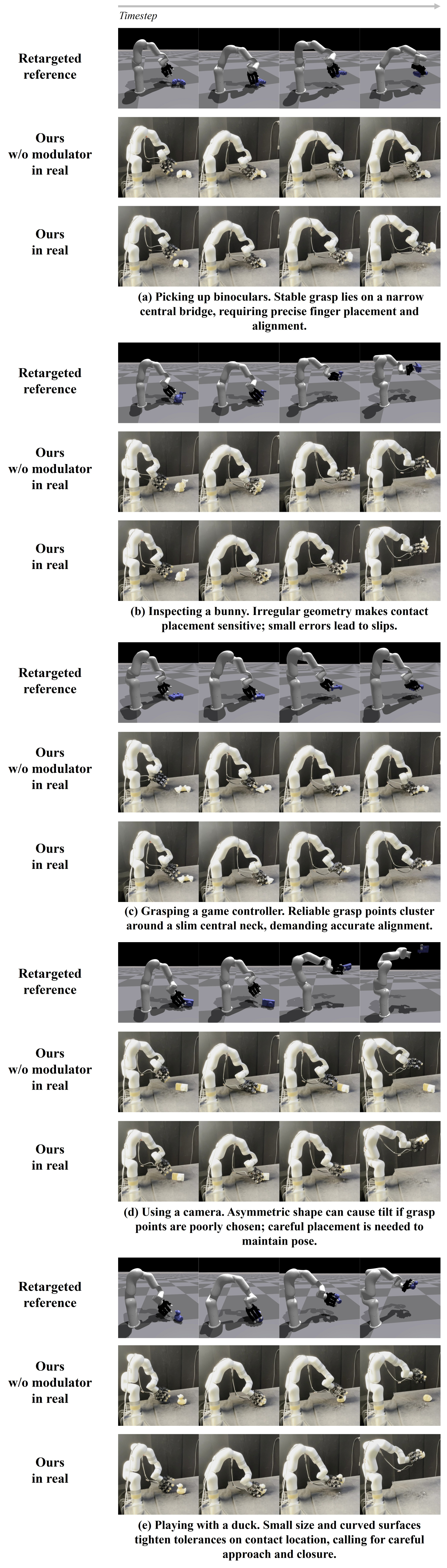}
  \vspace{-10pt}
  \caption{Additional qualitative results in real-world.}

  \vspace{-20pt}
  \label{fig:additional_real_experiments}
\end{figure}

Figure~\ref{fig:additional_real_experiments} shows additional qualitative comparisons on the unseen-trajectory set against the tracker-only baseline, highlighting AdaDexTrack’s robustness.

\section{Additional Analysis}
\label{sec:additional_analysis}

\subsection{Object Latent Smoothness}
\label{sec:latent_smoothness}

In our framework, object latent modulation is intended to act as a continuous interface for skill composition, rather than a discrete object identifier. The general tracker is conditioned on the object latent, and the modulator adjusts this latent online to recruit the most suitable behavior primitives for the current interaction phase. A key question, however, is whether such a mechanism is meaningful when the tracker is distilled from behavior cloning on a finite set of training objects. If the latent-to-action mapping were highly discontinuous, then interpolating in the latent space would not correspond to smooth interpolation between skills.

To examine this, we perform a Lipschitz-style perturbation analysis on the learned object latent. Specifically, we sample 200 reference observations from the unseen-trajectory set, and for each one we randomly select 5 timesteps, yielding 1{,}000 observations in total. For each observation, we perturb only the object latent by a small amount $\epsilon$, while keeping the robot state, object pose, and tracking reference fixed. We then measure the change in the tracker output using
\[
\rho = \frac{\|\Delta a\|_2}{\epsilon},
\]
where $\Delta a$ denotes the induced change in the predicted action.

\begin{figure}[H]
  \centering
  \vspace{-0.5em}
  \includegraphics[width=0.8\linewidth]{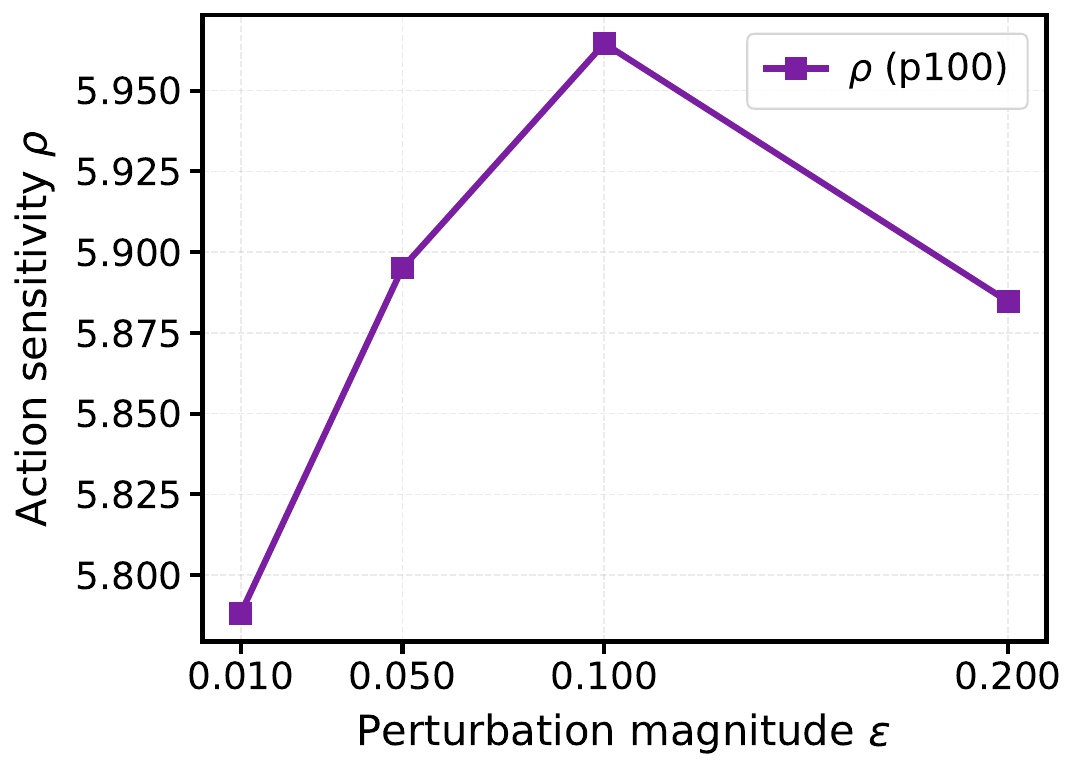}
  \caption{Lipschitz-style perturbation analysis of object latent modulation. The response ratio $\rho=\|\Delta a\|_2/\epsilon$ remains bounded under small latent perturbations, supporting a locally smooth latent-to-action mapping and meaningful interpolation between object anchors.}
  \label{fig:latent_smoothness}
  \vspace{-1em}
\end{figure}

As shown in Fig.~\ref{fig:latent_smoothness}, the response ratio $\rho$ remains bounded across samples and perturbation directions. This indicates that the tracker responds smoothly to local changes in the object latent, rather than switching erratically between unrelated behaviors. Such local smoothness is precisely the property needed for object latent modulation to function as a continuous control variable: nearby latent codes induce nearby actions, making interpolation between latent anchors meaningful at execution time.

This observation is consistent with the qualitative behavior shown in Fig.~\ref{fig:Modulation-object_latent} of the main paper, where the object latent acts like a phase-dependent ``knob'' for hand openness. The modulator can therefore use small latent adjustments to continuously recruit different grasping behaviors---e.g., preferring a more open hand during approach and a tighter hand during grasp---without changing the reference or relying on large action corrections. Overall, this analysis supports our claim that object latent modulation enables smooth interpolation and switching across the skill library learned by the distilled general tracker.

\end{document}